\documentclass[journal,transmag]{IEEEtran}
\usepackage{amssymb}
\usepackage{graphicx}        
\usepackage{multicol}        
\usepackage[bottom]{footmisc}
\usepackage{graphics}
\usepackage{amsmath,lipsum} 
\usepackage{mathrsfs}
\usepackage[lined,commentsnumbered,ruled]{algorithm2e}
\usepackage{xcolor}
\usepackage{comment}
\usepackage{subfigure}
\usepackage{float}
\usepackage{comment}
\usepackage{url}
\usepackage{mathtools}
\usepackage{stfloats}
\usepackage{ulem}
\usepackage{multirow}
\usepackage[colorlinks,citecolor=blue,urlcolor=blue,bookmarks=false,hypertexnames=true]{hyperref}

\usepackage[export]{adjustbox}
\usepackage{rotating}
\usepackage{booktabs, makecell, tabularx}
\usepackage{xcolor}

\newtheorem{myprob}{Problem}

\newcommand\numberthis{\addtocounter{equation}{1}\tag{\theequation}}

\begin{document}
	\title{Emergence of human oculomotor behavior from optimal control of a cable-driven biomimetic robotic eye}
	
	\author{\IEEEauthorblockN{Reza Javanmard Alitappeh\IEEEauthorrefmark{*,$\dagger$},
				Akhil John\IEEEauthorrefmark{$\dagger$},
				Bernardo Dias\IEEEauthorrefmark{$\dagger$}, 
				A. John van Opstal\IEEEauthorrefmark{+},\\ and
				Alexandre Bernardino\IEEEauthorrefmark{$\dagger$},~\IEEEmembership{Senior Member,~IEEE}}
			    \IEEEauthorblockA{\IEEEauthorrefmark{$\dagger$}Institute for Systems and Robotics, Instituto Superior T\'ecnico, ISR, Lisbon, Portugal}
			    \IEEEauthorblockA{\IEEEauthorrefmark{+}Dept. Biophysics, Donders Centre for Neuroscience, Radboud University, Nijmegen, The Netherlands}
		        \IEEEauthorblockA{\IEEEauthorrefmark{*}University of Science and Technology of Mazandaran, Behshahr, Iran}
		\thanks{
			Corresponding author: Reza Javanmard Alitappeh (email: Rezajavanmard64@gmail.com).}
	} 
	
	\IEEEtitleabstractindextext{%
		\begin{abstract}
			In human-robot interactions, eye movements play an important role in non-verbal communication. However, controlling the motions of a robotic eye that display similar performance as the human oculomotor system is still a major challenge. In this paper, we study how to control a realistic model of the human eye with a cable-driven actuation system that mimics the six degrees of freedom of the extra-ocular muscles. The biomimetic design introduces novel challenges to address, most notably the need to control the pretension on each individual muscle to prevent the loss of tension during motion, that would lead to cable slack and lack of control.
			We built a robotic prototype and developed a nonlinear simulator, for which we compared different approximation techniques to control its gaze behavior. In the first approach, we linearized the nonlinear model, using  a local derivative technique, and designed linear-quadratic optimal controllers to optimize a cost function that accounts for accuracy, energy expenditure, and movement duration. The second method uses a recurrent neural network that learns the nonlinear system dynamics from sample trajectories of the system, and a non-linear trajectory optimization solver that minimizes a similar cost function. For both approaches we developed a method to determine the required cable pretension at movement onset. We focused on the generation of rapid saccadic eye movements with fully unconstrained kinematics, and the generation of control signals for the six cables that simultaneously satisfied several dynamic optimization criteria. The model faithfully mimics the three-dimensional rotational kinematics and dynamics observed for human saccades. Interestingly, just like for the primate eye, the six cables organized themselves into the appropriate antagonistic muscle pairs that minimize the amount of muscle co-contraction. Our experimental results indicate that while both methods yielded similar results, the nonlinear method is more flexible for future improvements to the model, for which the calculations of the linearized model's position-dependent pretensions and local derivatives become particularly tedious.
		\end{abstract}
		%
		\begin{IEEEkeywords}
			oculomotor system; gaze control; recurrent neural network; degrees of freedom; main sequence kinematics; Listing's law; reciprocal innervation; muscle synergies; optimal control.
		\end{IEEEkeywords}
	} 
	
	\maketitle
	
	\IEEEdisplaynontitleabstractindextext
	
	\IEEEpeerreviewmaketitle
	
	\section{Introduction}
	
	With increasing utilization of robots in our daily lives, effective human-robot interaction (HRI) and communication is becoming an important challenge. Gaze plays an important role in social interactions,
	not only by signalling one’s attention to external events, but also to reflect attitudes, affects, or emotions \cite{argyle1973different}. This has spawned numerous studies of gaze behavior in human-robot interaction (see \cite{admoni2017social}\cite{kompatsiari2021s}\cite{Lohan2018}\cite{Duarte2018}, for a review), not only to enable robots to correctly respond to human gaze signals, but also (and more relevant to the present study) to equip robots with legible gaze behaviors when communicating with a human interlocutor. 
	
	Thus, an important objective for effective HRI systems is to develop biomimetic robotic eyes that can perform natural gaze behaviors  \cite{mcginn2020robots}\cite{de2020double}. 
	In addition, while the appropriate control of a biomimetic robotic eye can play a vital role in HRI systems, it can also be highly useful to better understand the human oculomotor system, including the vast variety of oculomotor disorders \cite{Ward2020}\cite{thurtell2007disorders}\cite{robinson2022}, the principles that guide active vision when scanning the environment \cite{peng2000active}\cite{Vasilyev2019}, or other applications that aim to model and understand sensorimotor behaviors that are similar to that of humans. 
	
	Most existing approaches to display realistic eye movements in robots either try to replicate pre-recorded psychophysical data from real human gaze shifts \cite{Duarte2018}\cite{kanda2002development}\cite{mohammad2010autonomous} or hard-code the biological control rules into the robot control system \cite{maini2008bioinspired}\cite{biamino2005mac}. Although these may seem viable approaches to design a biomimetic robotic eye, they are limited in several ways. First, the underlying principles that rule the emergence of the gaze kinematics are left uncovered. Thus, if the robot's mechanical structure does not perfectly match that of the human eye, the resulting behavior may not be ``optimal". Furthermore, if new movement tasks have to be implemented, new behavioral data will be needed, since it is not clear how to generalize the movement control principles from one task to another. Second, most existing models have only addressed relatively simple settings, and have, so far, neglected the planning and execution of the full dynamics that underlie the eye-movement trajectories. Moreover, typical robotic eyes are equipped with only 2 degrees-of-freedom (DOF), usually pan-tilt serial kinematics to control the azimuth and elevation of gaze, and cannot independently control ocular cyclotorsion. 
	%
	
	To address these limitations, in our recent work, we considered the generation of saccades (rapid eye movements) in a 3 DOF eye model \cite{Akhil2021}. The bio-inspired eye had three independent motors that were each coupled to an agonist-antagonistic cable pair. We proposed a feedforward optimal control that could reproduce the dynamics and kinematics of human saccadic eye movements in all directions, when the total cost comprising accuracy, energy expenditure, saccade duration, and total fixation force was minimized. In the present paper, we extend this work by making the eye model more biologically accurate and by removing the restriction of coupled agonist-antogonist cables. We designed a new scaled up biomimetic cable-driven eye prototype with six independent actuators that mimic the six extraocular muscles. A simulation of the prototype was used to investigate different techniques to find the optimal control policy. The proposed approaches would possibly lead to energy-efficient and more durable robotic systems, with more flexibility in replicating the complex repertoire of oculomotor behaviors exhibited by humans.

	\section{Background}
	\label{sec:background}
\begin{figure}
		\centering
		\includegraphics[width=0.7\columnwidth]{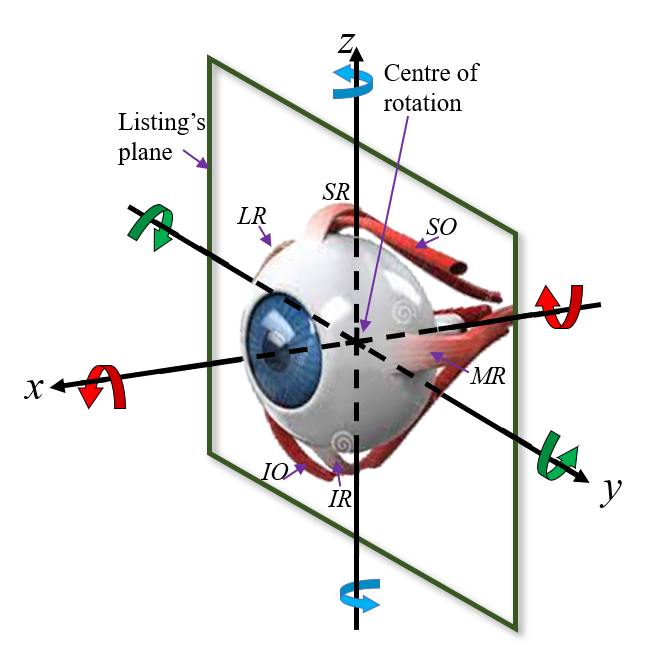}
		\caption{Representation of the right human eye with its six extra-ocular muscles. The muscles can rotate the eye in any three-dimensional orientation around its fixed center of rotation. The right-handed head-fixed reference frame of Listing (arrowheads) shows the three cardinal axes: horizontal ($y$), vertical ($z$) and torsional ($x$), respectively. For example, the horizontal recti rotate the eye mainly around the $z$-axis: lateral rectus, rightward (negative $z$); medial rectus, leftward (positive $z$). In Listing's frame of reference, Listing's plane coincides with the $yz$ plane and the $x$-axis points in the so-called primary direction}.
		\label{fig:eye}
	\end{figure}

	The eye is enclosed inside a conical cavity, where fats and other connective tissues restrict its translation \cite{robinson75}\cite{snell2013clinical}. Thus, the eye can effectively only rotate with three degrees of freedom. The human eye is actuated by six extraocular muscles that control its orientation (see Fig. \ref{fig:eye}). The muscles on the left and right of the right eye, the medial (MR) and lateral (LR) rectus, mostly rotate the eye horizontally. The top and bottom muscles, the superior (SR) and inferior (IR) rectus, along with the sideways pointing muscles, the superior (SO) and inferior (IO) oblique, allow for vertical and cyclo-torsional rotations. It is important to note that the pulling directions of these muscles are not completely independent of each other \cite{robinson75}. 
	As can be seen in Fig. \ref{fig:eye}, any eye orientation in 3D space can be obtained by a sequence of active head-fixed rotations around the $x, y$ and $z$ axes, respectively, specifying the \textit{torsional} ($\psi$), \textit{pitch} ($\varphi$) and \textit{yaw} ($\theta$) Euler angles. Each of these three rotational motions, are mainly driven by actuating a pair of extra-ocular muscles in an approximate agonist-antagonist configuration, where each muscle pulls the eye in a different direction.
	
	\subsection{Saccade kinematics}
	
	Typically, humans make about 3-4 saccades per second to scan the visual environment \cite{robinson2022}\cite{van2016auditory} and these are constrained by Donders' law, which restricts the rotational degrees-of freedom of the eyes from three to two \cite{donders1864anomalies}. \textit{Donders' law} states that any eye orientation has a unique cyclotorsional angle, regardless of the path followed by the eye to reach that orientation. In other words: $\psi=f(\theta,\phi)$, where $f(\theta,\phi)$ depends on the motor system \textit{(eye, or head)} and motor task (e.g., vergence, vestibular, etc.).  It has been argued that Donders' law avoids the problems associated with the non-commutativity of rotations in three dimensions, which becomes especially important when planning sequences of eye and head movements \cite{tweed87}\cite{hepp1990Listing}\cite{opstal91}. 
	\textit{Listing's law} is a further specification of Donders' law. It provides an extra restriction on the eye's cyclotorsional angle for the special condition with the head upright and still, and the eye looking at infinity. With the Euler angles as defined above, Listing's law states: $\tan(\psi/2) = \tan(\varphi/2)\cdot \tan(\theta/2)$. In the axis-angle formulation of rotations, $R(\psi, \varphi, \theta) \equiv \{\hat{\bf{n}}, \rho\}$, where $\hat{\bf{n}}$ is a unit vector and $\rho$ is the angle. The neuroscience literature typically uses the Euler-Rodrigues rotation vector ($\mathbf{r}$) to represent 3D eye orientations, where $\mathbf{r} \equiv \tan(\rho/2)\cdot \hat{\mathbf{n}}$ \cite{tweed87}\cite{opstal91}\cite{HASLWANTER1995}\cite{van1996role}\cite{van2002gaze}. In this representation, Listing's law constrains all eye orientations to a plane, which in the laboratory frame ($\mathbf{r}=\mathbf{0}$ is straight ahead) is described as $r_x = a\cdot r_y + b\cdot r_z$. This plane is denoted as Listing's plane. A change of coordinates can express rotations in a new reference frame, where Listing's plane is aligned with the $(yz)$ plane, i.e., ${r}_x = 0$. Now, $\mathbf{r}=\mathbf{0}$ is the physiologically defined primary position (Fig.~\ref{fig:eye}). Note that Listing's law not only holds during steady eye fixations, but also during smooth-pursuit eye movements and rapid saccadic eye movements. The law does not hold during eye-head coordination, static head tilts, vestibular and optokinetic stimulation, or for disjunctive vergence eye movements to fixate nearby targets \cite{tweed95}. For those movements, Donders' law applies in a task-specific manner.
	\subsection{Saccade dynamics}
\begin{figure*}[t]
		\centering
		\subfigure[]{\includegraphics[width=0.38\columnwidth]{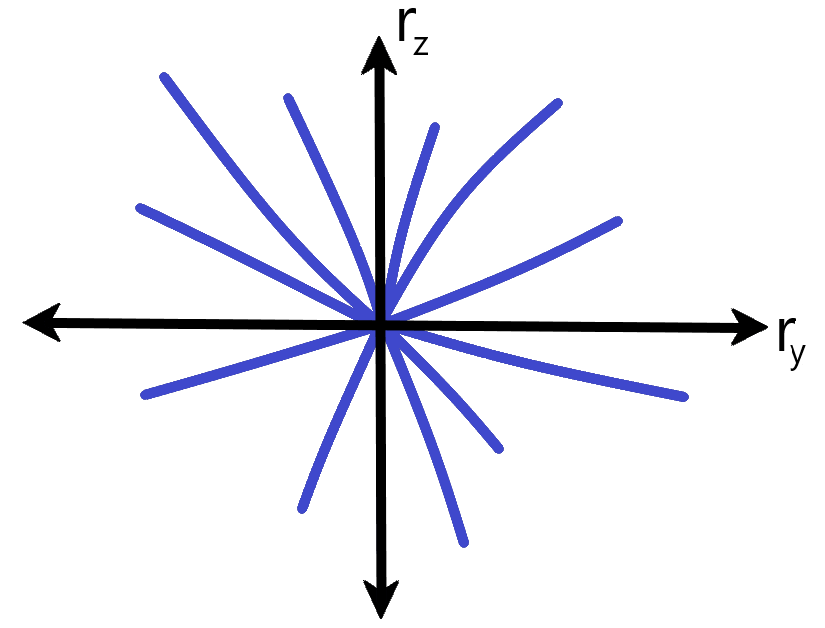}}~~
		\subfigure[]{\includegraphics[width=0.38\columnwidth]{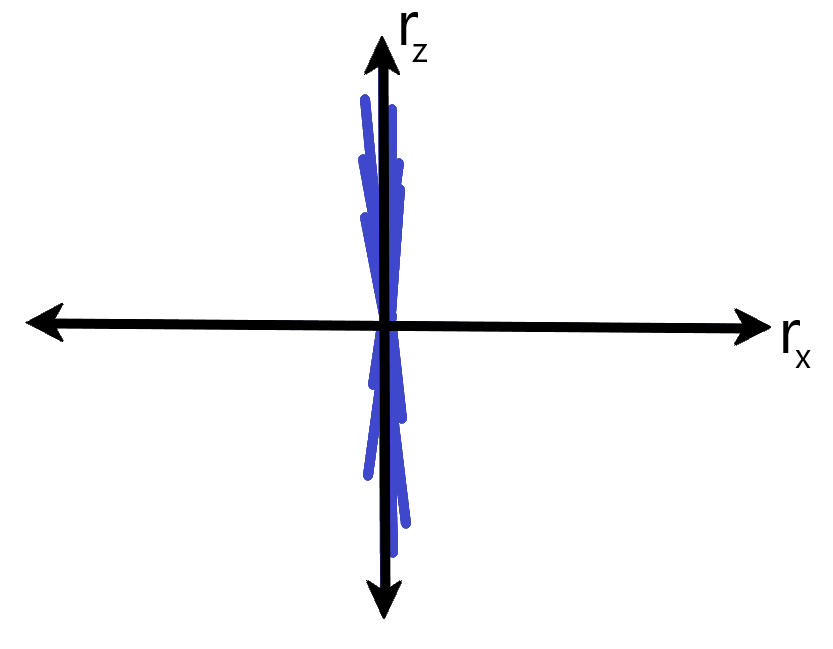}}~~
		\subfigure[]{\includegraphics[width=0.38\columnwidth]{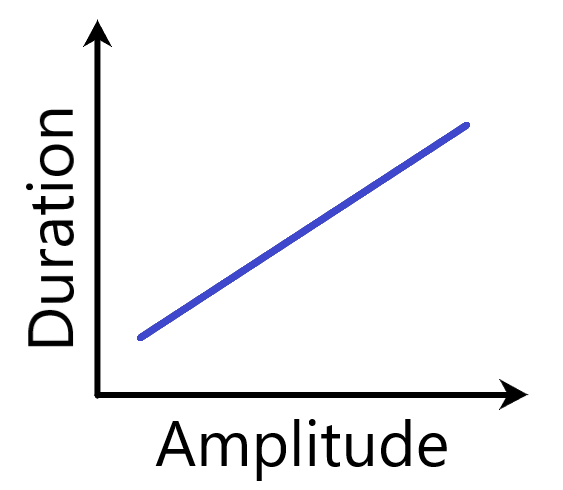}}~~
		\subfigure[]{\includegraphics[width=0.38\columnwidth]{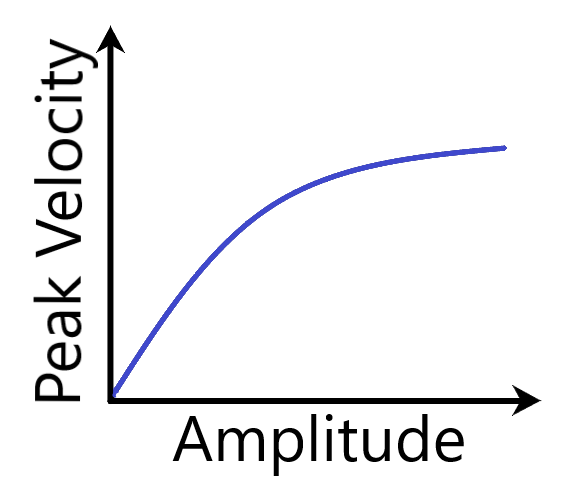}}~~
		\subfigure[]{\includegraphics[width=0.38\columnwidth]{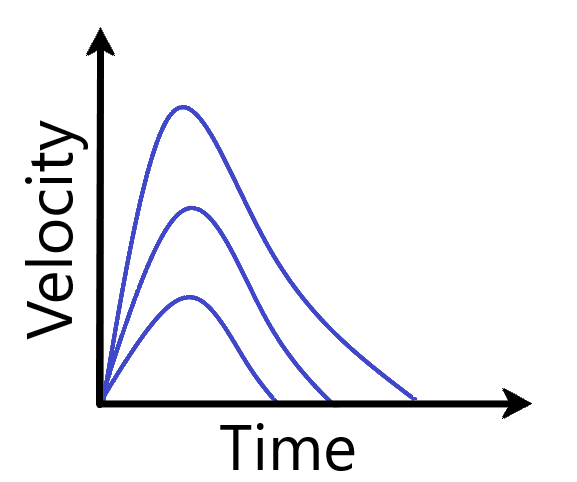}}	
		\caption{Schematic illustration of saccade kinematics (a,b) and dynamic (c,d,e) properties. Panels (a) and (b) show the $(r_y, r_z)$ and $(r_x, r_z)$ views of the thin plane formed in 3D when eye orientations during saccades are represented by rotation vectors (Listings' Law: $r_x = 0$). Note nearly straight trajectories in all directions (a). Panel (c) shows the affine relationship between the saccade duration and amplitude (longer durations for larger saccades). Panel (d) shows the saturation of peak eye-velocities with increasing saccade amplitudes. The increasing skewness of saccade velocity profiles with amplitude, and the nearly constant eye-acceleration times, are shown in panel (e).}
		\label{fig:eye_movement_properties}
	\end{figure*}

	A further important property of saccadic eye movements is their nonlinear dynamics, described by the so-called \textit{main sequence}. In humans, saccade peak velocity saturates at large amplitudes, which follows from the affine increase of saccade duration with amplitude: $D=a\cdot A+b$. Indeed, since normal sac\-cade velo\-city profiles have single-peaked 'triangular' shapes, their peak, $V_{pk}$, relates to their width ($D$) and amplitude by $V_{peak}\cdot D = 2A$, leading to $V_{pk} \sim 2A/(aA + b)$, which saturates at $2/a$ deg/s. In addition, saccades reach their peak velocity after an approximately fixed acceleration phase of about 20-25 ms \cite{Opst87}\cite{yang2001eye}, regardless their amplitude. This causes the skewness of the velocity profile to increase from near zero (symmetric) for small saccades, to positively skewed profiles for large saccades \cite{Opst87}. Schematic plots for these properties are shown in  (Fig. \ref{fig:eye_movement_properties}).

	Finally, behavioral experiments have shown that oblique saccade trajectories are approximately straight \cite{Gisb85}. As a consequence, the horizontal and vertical component velocities are scaled versions of each other (i.e., they are synchronized), resulting in a significant stretching of the smaller component when it participates in an oblique saccade. For example, a purely horizontal 10 deg saccade can have a peak velocity of 300 deg/s and a duration of 50 ms. However, when it is part of a 20 deg oblique saccade, made at an angle of 60 deg with the horizontal, the peak velocity of the 10 deg horizontal component will be reduced to about 150 deg/s ($\cos(60^o)\cdot 300$) and its duration will have increased to about 80 ms (corresponding to the duration of a 20 deg saccade).
	
	%
	\subsection{Cable-driven robots and cable tension}
	Cables transmit force by applying tension and must therefore remain under tension at all times. To ensure proper function of a cable-driven robotic eye and to prevent actuator backlash, the six cables are pretensioned \cite{Kosari2013ControlAT}. However, the amount of cable pretension has to be tuned with care: excessive pretension causes more friction, which leads to faster wear and tear and shorter lifetime \cite{CableElasticity}. Insufficient pretension causes slack and deficiency in control \cite{Kosari2013ControlAT}. Similarly, recordings from primate oculomotor neurons have indicated that with the eye at rest in the primary position, about 65\% of the neural population is recruited \cite{robinson1972}. Effectively, this means that the eye is kept under continuous pretension and that the neural control for eye movements modulates the relative innervations of agonist and antagonist muscles by changes in firing rates to enable high-precision angular control of the eye. 
	\section{Related work}
	\label{sec:relatedwork}
	
	Even though saccadic eye movements have been studied for decades, they were mostly restricted to 1D (typically horizontal). The use of robotic eye models for implementation and understanding humanlike eye movements has a been fairly recent development \cite{maini2008bioinspired}\cite{biamino2005mac}.  In \cite{maini2008bioinspired} a neurophysiologically inspired model of combined saccadic eye-head movements in a robotic eye-head system was implemented, in which each of the two eyes had two degrees-of-freedom (pan and tilt). The system could successfully replicate the velocity profiles of human eye-head gaze shifts\cite{goossens97}, although experiments were performed exclusively along a single dimension, i.e. either horizontal or vertical saccades. In \cite{biamino2005mac} a tendon-driven mechanical eye (the MacEye) was designed to comply with Listing's law (see section~\ref{sec:background}), by appropriate routing of the cables and precisely calculated insertion points. The orientations of the eye complied with Listing's law, but the dynamics of the eye's behavior were not presented. Although a hardwired implementation of Listing's law seems an interesting engineering solution, it will only be valid for head-restrained eye movements with the head upright and for gazing at far objects. Thus, the MacEye system lacked the true mechanical degrees of freedom that would allow it to generate natural movements of the eyes with three degrees of freedom.
	
	A related study implemented an open-loop neural controller with a local adaptation technique to control an eye-head robotic model \cite{saeb2011learning}. The model followed the nonlinear behavior of the oculomotor system, but in contrast to the present study it utilized a high-level signal for the system without considering the complexity of 6 independent motors to control.
	
	In our recent work, we tested a feedforward open-loop optimal control strategy \cite{shadmehr2012} based on a linear approximation of a nonlinear biomimetic robotic eye to reproduce most of the dynamics and kinematics of human saccadic eye movements in a 3 DOF unconstrained robotic eye \cite{Akhil2021}. The biomimetic eye had three independent motors that were each coupled to an agonist-antagonistic cable pair. Linear approximation for the open-loop controller was done using systems identification of the robotic physics-based simulator. The successful cost function for the optimal control that enabled the reproduction of most human eye-movement characteristics minimized the joint costs for response accuracy, total energy expenditure, saccade duration, and total force exerted on the eye during eccentric fixations.
	%
	%
	
	\section{Problem Statement}
	\label{sec:problemstatement}
	In this paper we extend the state of the art by developing a 6 DOF biomimetic artificial eye with cables mimicking the human extraocular muscles, and propose control laws that replicate the most important characteristics of human saccadic eye movements. This brings new challenges to the control, like setting the appropriate muscle pre-tensions that impact on the highly redundant muscular system in the generation of near-optimal eye movements. For this new system, we developed control laws that result to replicate most kinematic and dynamic characteristics of human saccades, including the emergence of an agonist-antagonist organization of distinct muscle pairs from basic optimality criteria.
	Similarly to our previous work \cite{Akhil2021} we take a model-based stance to the control problem, which is here divided into two steps:
	deriving forward models of the mechanical eye system, suitable for control design (\textit{Problem 1}), and computing open-loop optimal control trajectories for the modelled system (\textit{Problem 2}).
	
	\begin{myprob} [Approximating the forward dynamic model]
		\label{problem1}
		\textnormal{}
		we analyzed two approaches to model the physics-based dynamic equations of the cable-driven robotic eye system of Fig.~\ref{fig:eyeprototype}, which we here denote by a \textit{linear and a nonlinear} approximation, respectively. We represent the motor angles as $\mathbf{u} \equiv [ u_{IR}, u_{MR}, u_{SR}, u_{LR}, u_{IO}, u_{SO}]^T$ and the 6 DOF eye state was composed of its 3D orientation and angular velocity: $\mathbf{x}\equiv [r_x, r_y, r_z, \omega_x, \omega_y, \omega_z ]^T$. In the \textit{linearized} approach, we computed a linearized approximation of the system equations; in the \textit{nonlinear} method, the eye-movement behavior was learned by applying the nonlinear system's output to a recurrent neural network. In both approaches, the next state was predicted from a given control input $\mathbf{u}_t$ for the 6 motors and the current state $\mathbf{x}_t$, where $t$ indicates discrete time. Thus, for both approaches, the model  could be defined by the evolution of the state, under the action of the command:
		\[
		\mathbf{x}_{t+1}=\mathbf{f}(\mathbf{x}_t, \mathbf{u}_t)
		\]
		
	\end{myprob}
	\begin{myprob} [Trajectory optimization]
		\label{problem2}
		{
			A human-like saccadic eye movement is specified by the optimal trajectory $\mathbf{u}^*_{0:T}$), which brings the eye from the initial state $\mathbf{x}_0$ to its final state $\mathbf{x}_T$ in a finite time-interval $t \in \left[ 0, T \right]$. 
			
			If we consider $\mathbf{u}_{0:T}$ as the sequence of input motor commands $\mathbf{u}_{0:T}=[\mathbf{u}_{0}, \mathbf{u}_1, \cdots, \mathbf{u}_T]$, the problem then is to optimize the input motor commands, such that they lead to the optimal trajectory at minimum cost. This cost is typically composed of a linear combination of partial costs on the properties of the trajectory, e.g. \textit{duration}, \textit{accuracy}, \textit{energy}.
			The minimization can thus be written as: 
			\begin{align*}
				&\min_{0 \leq T \leq T_{\max} \; } \bigg( \min_{\mathbf{u}_{0:T}} \; \sum_{t=0}^{T} C(\mathbf{x}_t,\mathbf{u}_t, T)\bigg),~\\
				&\mbox{subject to} :~ \mathbf{x}_{t+1}=f(\mathbf{x}_{t},\mathbf{u}_{t}), \\
				& \qquad \qquad \quad \mathbf{u}_t \in {\cal U}
				& \numberthis \label{eq_probdef2}  
			\end{align*}
			where, $C()$ indicates the cost of input state $\mathbf{x}$ and motor command $\mathbf{u}$ for a given duration $T$, $T_{\max}$  is a bound on the possible values of $T$, and ${\cal U}$ is the set of feasible commands. The optimization process is typically organized in an inner optimization of the motor commands $\mathbf{u}$ for a fixed time horizon $T$, and an outer loop that optimizes $T$ in a certain range $0 \leq T \leq T_{max}$.
			Detail about the components in the cost function is provided in Sec. \ref{sec:OptimalControl}.}
	\end{myprob}
	%
	
	\section{Design of our Cable-driven Robotic Eye}
	\label{sec:cabledrivensye}
	\begin{figure}[t]
		\centering
		\subfigure{\includegraphics[scale = 0.45]{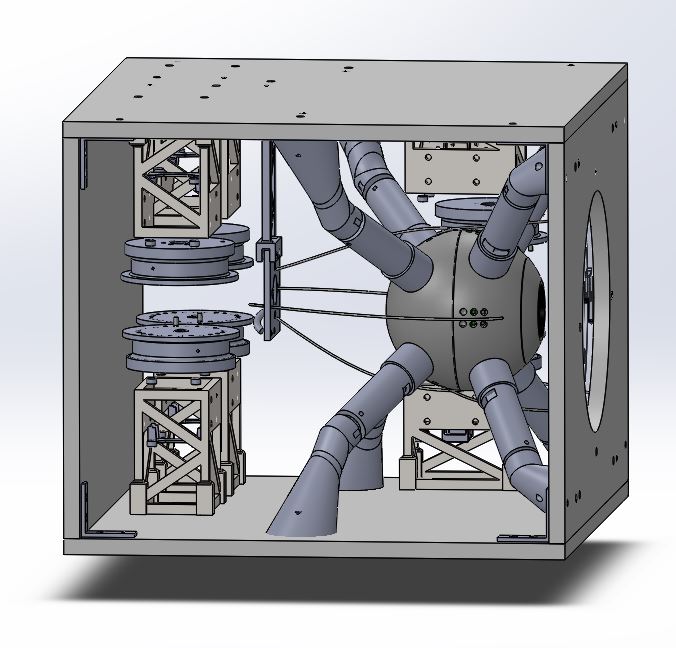}}\\
		\subfigure{\includegraphics[scale = 0.1]{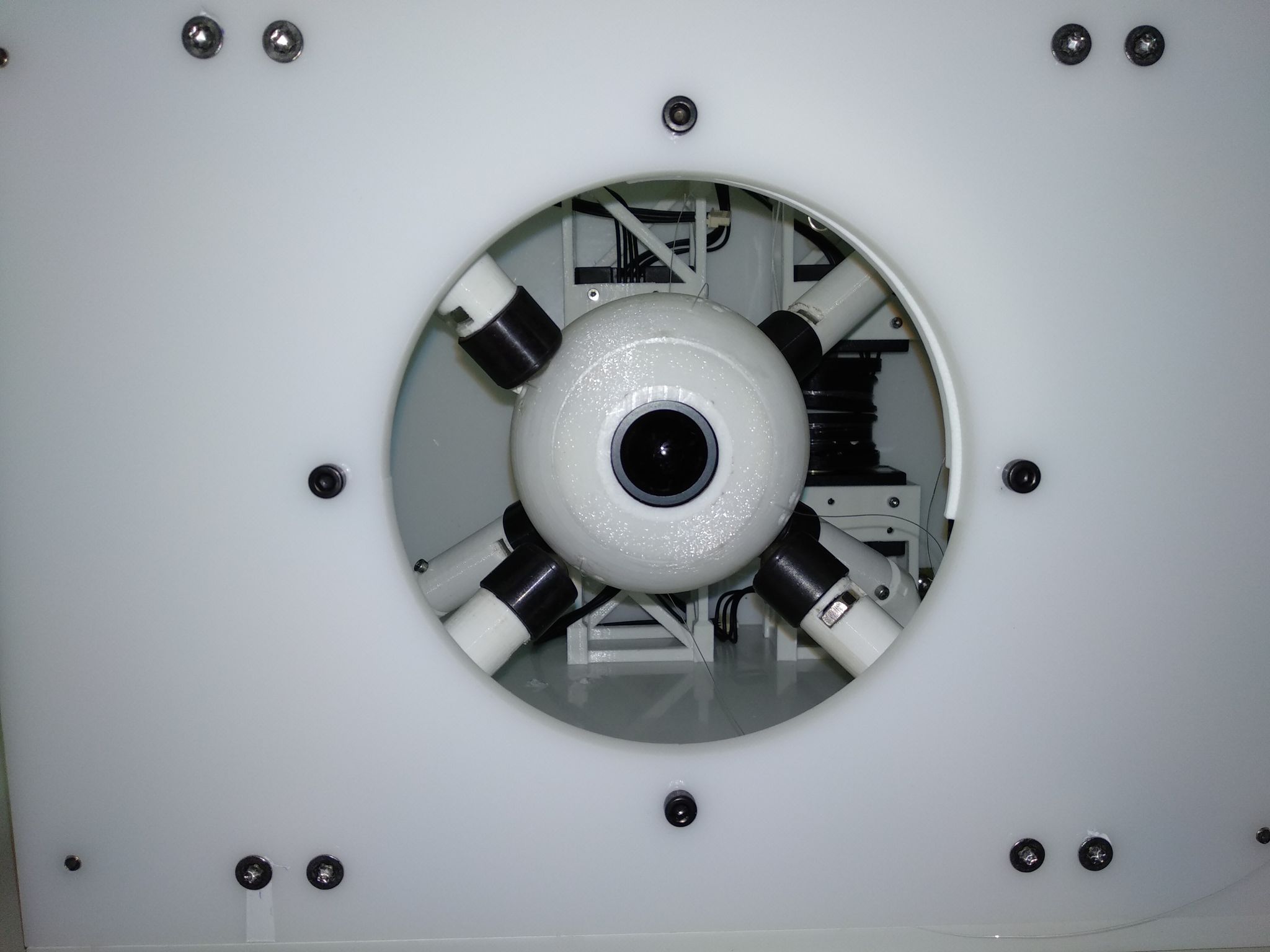}}
		\caption{Top: Schematic 3D side view of our biomimetic human eye prototype, with six motors (five of its spindles are visible) independently controlling the six cables (four are visible) connected to the eyeball. The eye is kept in place by the eight external arms to only allow three-dimensional rotations around its fixed center. The ball-contacts of these arms on the eye ball (not visible) provide a dynamic frictional force that increase the total damping of the system.  The six cables, controlled by motors, correspond to the extra-ocular muscles of Fig.~\ref{fig:eye}. Bottom: Front view of the actual mechanical prototype showing the eye with the camera.}
		\label{fig:eyeprototype}
	\end{figure}
	\begin{figure}[t]
		\centering
		\includegraphics[scale = 0.35]{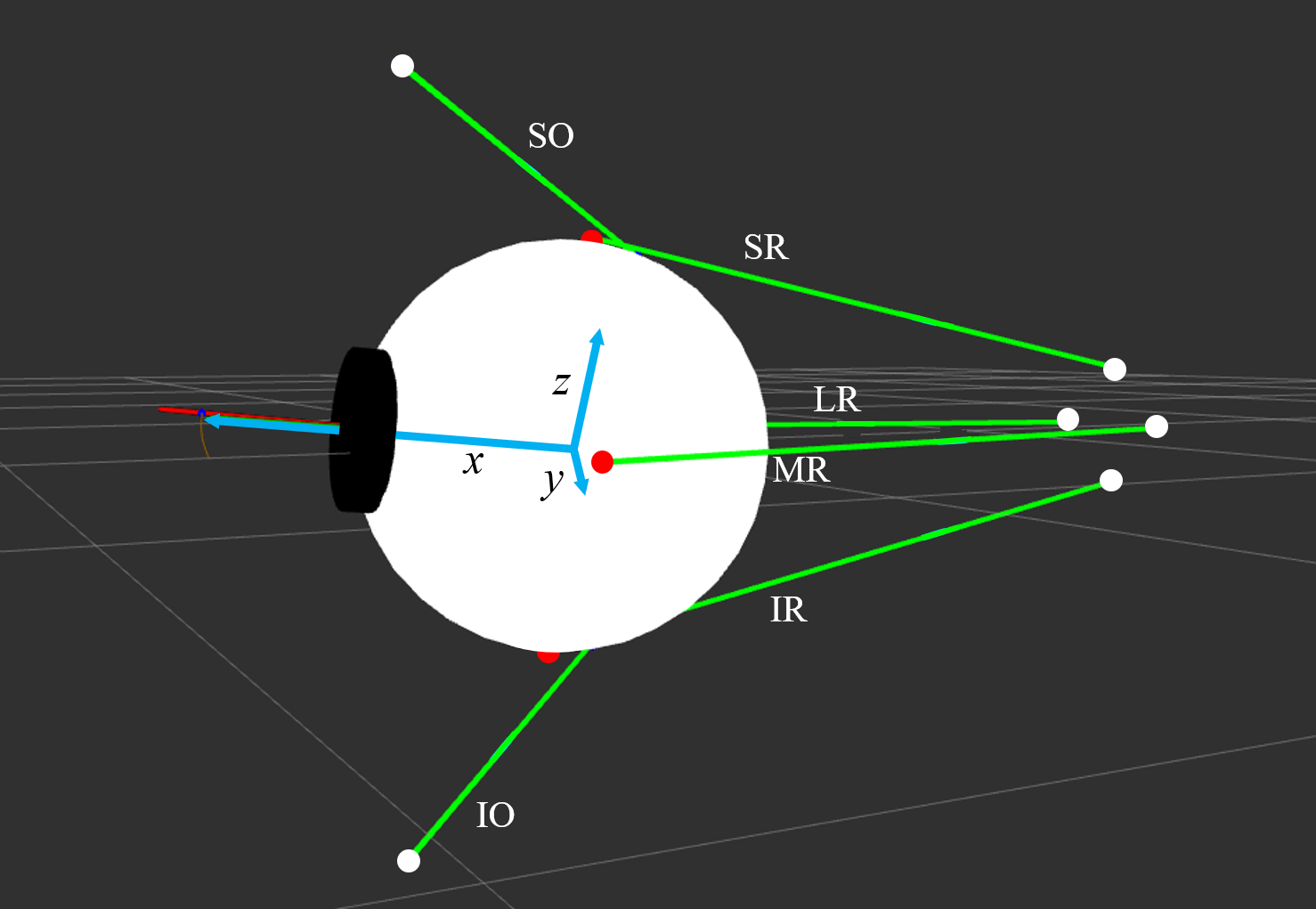}
		\caption{Visualizaton of the EOM (Extra-Ocular Muscles) setup for the simulator of the right eye deviated from the straight-ahead origin; light blue laboratory frame) to the right (red line), where red and white dots indicate insertion points on the eye and on the head, respectively. Note that cable pulling directions are not symmetric about the center of the eye, because the insertion points on the head at the back are shifted medially in the $+y$ direction. As a result, the length of the MR muscle is shorter than the LR when the eye is in the primary position.}
		\label{fig:graphical}
	\end{figure}

	Figure~\ref{fig:eyeprototype} shows the 3D model and the constructed mechanical prototype of our system. Like the human eye, the robotic eye will rotate around its fixed center whenever the six elastic cables, which represent the extra-ocular muscles, exert a net torque. The cables are inserted on the globe at similar contact points as on the human eye, to allow it to rotate with 3 DOF. Each cable is controlled by its own motor that rotates at a controlled speed, thus pulling the cable around its spindle to exert a torque on the eye. Because the pulling directions of the cables vary with the orientation of the eye, and each muscle can only pull (not push), the total system (as described below in more detail) is \textit{nonlinear} \cite{Akhil2021}, and to rotate the robotic eye in the same way as the human eye during rapid saccades (i.e., accurate and at high speeds) becomes a highly nontrivial problem. 
	
	In order to study the properties and behavior of this biomimetic oculomotor system we developed an eye simulator. The eye was modeled as a sphere with a fixed center, subject to Newton-Euler’s rigid body equation of angular motion \cite{multibody} (Eq. \ref{eq:eye_acceleration1}) actuated by the six cable-driven actuators:
	\begin{equation} \label{eq:eye_acceleration1}
		\prescript{e}{}{\boldsymbol{\alpha}_{h,e}} = \prescript{}{e}{\textbf{I}_e}^{-1}(\prescript{}{e}{\boldsymbol{\tau}_{net}(\boldsymbol{x},\boldsymbol{u})}-\prescript{e}{}{\boldsymbol{\omega}_{h,e}} \times \prescript{}{e}{\textbf{I}_e}\prescript{e}{}{\boldsymbol{\omega}_{h,e}})
	\end{equation}
	where $\prescript{e}{}{\boldsymbol{\alpha}_{h,e}}$ is the angular acceleration of the eye with respect to the head frame, expressed in the eye frame; $\prescript{}{e}{\textbf{I}_e}$ represents the inertia tensor of the eye model; $\boldsymbol{u}$ is the motor configuration of the 6 motors; $\prescript{e}{}{\boldsymbol{\omega}_{h,e}}$ is the angular velocity of the eye, the symbol $\times$ denotes the vector (cross) product, and $\prescript{}{e}{\boldsymbol{\tau}_{net}}$ is the net torque exerted on the eye. The net torque depends on the dynamic friction and elasticity torques, $\prescript{}{e}{\boldsymbol{\tau}_d}$ and $\prescript{}{e}{ \boldsymbol{\tau}_k}$, respectively:
	\begin{equation} \label{full eye torque}
		\prescript{}{e}{ \boldsymbol{\tau}_{net}}=  \prescript{}{e}{ \boldsymbol{\tau}_k} +\prescript{}{e}{\boldsymbol{\tau}_d} = \sum_{m = 1}^{6} \prescript{}{e}{ \boldsymbol{\tau}_m} - \textbf{D}_{eye}\prescript{e}{}{\boldsymbol{\omega}_{h,e}}
	\end{equation}
	where $\textbf{D}_{eye}$ quantifies the eye's damping matrix, subscript $m$ is the motor index, and
	\begin{equation}
		\prescript{}{e}{\boldsymbol{\tau}_m} = \prescript{e}{}{\textbf{Q}_{m}} \times \prescript{}{e}{\textbf{f}_m}\label{muscle torque individual}
	\end{equation}
	is the torque exerted by each muscle, $\prescript{e}{}{\textbf{Q}_{m}}$ is each muscle's insertion point on the eye in the eye frame, and $\prescript{}{e}{\textbf{f}_m}$ is the force applied by each muscle on the eye ball, that depends on the current state $\mathbf{x}_t$, the goal state, $\mathbf{x}_G$, and the control input $\mathbf{u}_t$. 
	
	To simplify the modeling in the simulations, the elastic cables were approximated by linear elastic springs. So the elastic force applied on the eye by each cable ($\prescript{}{e}{\textbf{f}_m}$) depends on its length ($l_m$), which is determined by the sum of the cable length wound on the motor spindle and the length between the head-fixed routing point of the cables (represented by the white points in Fig.~ \ref{fig:graphical}) and the final eye-fixed contact point on the eye (red points).
	The length of the cable for each muscle ($l_{m}$) varies with the rotation of the motors ($\mathbf{u}$) and orientation of the eye (from state $\boldsymbol{x}$; we omit time index t, for clarity), 
	\begin{equation}
		l_m(\mathbf{x},\mathbf{u}) = \left\lVert \prescript{h}{}{\textbf{P}_{eye,m}}(\mathbf{x}) - \prescript{h}{}{\textbf{P}_{head,m}} \right\rVert + r\cdot u_m\label{eq:length}
	\end{equation}
	where $\prescript{h}{}{\textbf{P}_{eye}}$ and $\prescript{h}{}{\textbf{P}_{head}}$ are the insertion points of cable m in the head reference frame, $r$ is the spindle's radius, and $u_m$ is the rotation angle of the spindle for cable $m \in \{IR, MR, SR, LR, IO, SO\}$. This leads to a dynamic elastic force that is determined by Hooke's law \cite{rychlewski1984hooke}:
	\begin{equation}
		\prescript{}{e}{\textbf{f}_m} = \frac{k}{l_{0m}}(l_m(\mathbf{x},\mathbf{u})-l_{0m})\prescript{}{e}{\boldsymbol{\Vec{\phi}_m}}\label{eq:force}
	\end{equation}
	where $k$ is a constant depending on the material and thickness of the cables (taken 20 N for all cables) and $l_{0m}$ is the length of cable $m$ when it is not stretched. $\prescript{}{e}{\boldsymbol{\Vec{\phi}_m}}$ is the unit vector in the direction in which the force is applied on the eye.
	
	Also, the system’s inertia, stiffness and damping parameters 
	were defined to closely replicate the time constants and overdamped characteristics of the human eye (see \cite{robinson2022}).
	
	Note that elastic cables can only apply pulling forces unlike ideal linear springs. To implement this constraint in our simulator, the force was set to zero as soon as it went negative. When this happens, the cable becomes loose and no longer applies tension, a phenomenon known as \textit{slack} (see \ref{subsec::pretension} Controlling pretension).
	
	Note also that since the elasticity constant, $k$, was taken identical for all cables, the effective stiffness varies for movements in different directions. For example, for horizontal movements, the elastic forces are primarily delivered by LR and MR, but vertical movements involve the interaction of SR,IR, SO and IO.
	
	
	\section{Approximations of the system dynamics}
	%
	%
	\subsection{The linearized model}
	\label{sec:linearizedmodel}
	We performed a local derivative-based linearization on the non-linear system dynamics (Eqns.~\ref{eq:eye_acceleration1}-\ref{eq:force}), by applying an infinitesimal perturbation method around an equilibrium point \cite{variation_linear1}.
	To that end, we applied the Lie Group \cite{iserles2000lie} transformation of the state variables to create a local linear state from the global equilibrium state, to which the perturbation was applied. Exponential mapping then yields the orientation and angular velocity in the local state, which can subsequently be transformed into the global (nonlinear) state.
	To illustrate this procedure, consider the following set of differential equations that represent the system's nonlinear dynamics:
	\begin{equation}
		\begin{gathered}\label{se} 
			\Dot{\prescript{w}{}{\textbf{R}_{e}}} = \prescript{w}{}{\textbf{R}_{e}}\prescript{e}{}{\boldsymbol{\omega}_{w,e}^{\wedge}}\\
			\prescript{e}{}{\Dot{\boldsymbol{\omega}}_{w,e}} =\prescript{}{e}{I_{e}}^{-1}(\prescript{}{e}{\boldsymbol{\tau}_{net}} - \prescript{e}{}{\boldsymbol{\omega}_{w,e}} \times \prescript{}{e}{I_{e}}\prescript{e}{}{\boldsymbol{\omega}_{w,e}}) \\
		\end{gathered}
	\end{equation}
	where $\prescript{w}{}{\textbf{R}_{e}}$ is the rotation matrix that converts from eye reference frame into the world reference frame and for a vector $\boldsymbol{v}=(x;y;z)$, $\boldsymbol{v}^{\wedge}$ (read vee hat) is defined as a 3x3 skew-symmetric matrix
	\begin{equation}
		\begin{gathered}\label{skkk} 
			\boldsymbol{v}^{\wedge}=\left[\begin{array}{l}
				x \\
				y \\
				z
			\end{array}\right]^{\wedge}:=\left[\begin{array}{rrr}
				0 & -z & y \\
				z & 0 & -x \\
				-y & x & 0
			\end{array}\right]
		\end{gathered}
	\end{equation}
	
	Let us define an equilibrium point as
	\begin{equation} 
		\bar{\textbf{x}} = \left\{ {}^w\bar{\textbf{R}}_e, {}^\omega\bar{\boldsymbol{\omega}}_{w,e}, \right\}. 
	\end{equation}
	
	A local state around the equilibrium point $\bar{\textbf{x}}$ is defined as 
	\begin{equation} 
		\tilde{\textbf{x}} = \left\{ {}^w\tilde{\textbf{R}}_e, {}^e\tilde{\boldsymbol{\omega}}_{w,e}, \right\} 
	\end{equation}
	with 
	\begin{equation}
		\begin{gathered}
			{}^w\tilde{\textbf{R}}_e = {}^w \bar{\textbf{R}}_e^T {}^w \textbf{R}_e \\
			{}^e\tilde{\boldsymbol{\omega}}_{w,e} = {}^e \boldsymbol{\omega}_{w,e} -{}^e\bar{\boldsymbol{\omega}}_{w,e} \\
		\end{gathered} 
	\end{equation}
	where $\tilde{\textbf{R}}$ is the rotation between the equilibrium orientation $\bar{\textbf{R}}$ and the actual orientation $\textbf{R}$. Using exponential notation, $\tilde{\textbf{R}}$ can be represented by the skew-symmetric matrix of a rotational perturbation $\boldsymbol{\eta}(t)$, where $t$ denotes time.
	\begin{equation}
		\tilde{\textbf{R}}(t) = \exp(\boldsymbol{\eta}^{\wedge}) \label{perturbation}
	\end{equation}
	
	Adapted from \cite{variation_linear1}, an infinitesimal variation, with respect to a reference $\bar{\textbf{R}}(t)$ $\in SO(3)$ is given by
	\begin{equation}
		\delta{\textbf{R}}(t)=\left.\frac{d}{d \epsilon}\right|_{\epsilon=0} \bar{\textbf{R}}(t) \exp (\epsilon \boldsymbol{\eta}^{\wedge})=\bar{\textbf{R}}(t) \boldsymbol{\eta}^{\wedge}(t)
	\end{equation}
	
	where $\epsilon$ is a small rotation around an axis $\boldsymbol{\eta}$.
	
	\par From \cite{yu}, the corresponding infinitesimal change of the eye's angular velocity is given by:
	
	\begin{equation}
		\delta \boldsymbol{\omega}(t)=\boldsymbol{\omega}^{\wedge}(t) \boldsymbol{\eta}(t)+\dot{\boldsymbol{\eta}}(t) \label{delta_o}
	\end{equation}
	
	The local state can therefore be represented in local coordinates and matrix form as:
	\begin{equation}
		\dot{\boldsymbol\xi} = \frac{d}{d t}\left[\begin{array}{c}
			\boldsymbol{\eta}_e \\ 
			\delta \boldsymbol{\omega}_e 
		\end{array}\right]=\left[\begin{array}{c}
			-\boldsymbol{\omega}_e^{\wedge}\boldsymbol{\eta}_e + \delta \boldsymbol{\omega}_e \\
			
			{}_{e}{I}_e^{-1} \left( {}_e \boldsymbol{\tau}_e - \boldsymbol{\omega}_{e} \times {}_{e}{I}_e \boldsymbol{\omega}_{e}\right)\\
			
		\end{array}\right]
	\end{equation}
	
	The linearized state-space model is thus obtained as:
	
	\begin{equation}\label{ss}
		\dot{\boldsymbol{\xi}} = \textbf{A}\boldsymbol{\xi}+\textbf{B}\delta\textbf{u}
	\end{equation}
	More detail of this linearization can be found in \cite{Bernardo2021} for the interested reader.
	This linear model proved to accurately approximate the non-linear system for small perturbations around the linearization point. The difference
	between the linearized and original nonlinear system increased with the eye's gaze eccentricity when the perturbation was applied to the eye's orientation state. 
	In contrast, a perturbation of its velocity led to smaller deviations from the non-linear system (see Table \ref{linear_err}). 
	
	\begin{table}[htb!]
		\centering
		\caption{maximum perturbation (in deg) for which the overall accuracy error remains below 5 \%.}
		\label{linear_err}
		\begin{tabular}{||c | c | c ||} 
			\hline
			Perturbation & at Origin &  at 30 deg eccentricity\\ 
			\hline
			$\delta R_{eye}$ &  $7.85^\circ$ & $5.89^\circ$\\
			\hline
			$\delta \omega_{eye}$ & No max & No max\\
			\hline
		\end{tabular}
	\end{table}
	
	The drop in accuracy with larger perturbations has a direct impact on the accuracy of planned saccades, since an open-loop control strategy requires a good model of the system. This means that a saccade planned and executed with the linearized local state dynamics might not end in a stable equilibrium in the global state. Thus, the system has to be forced into an equilibrium configuration for that global state after each saccade. This was achieved by the optimization discussed later in Section \ref{subsec::pretension}.
	
	%
	\subsection{The nonlinear NARX model}
	\label{subsec:NARX}
	An alternative approach in learning the optimal control in complex robotics applications has been the application of machine-learning techniques, including  deep learning and reinforcement learning 
	\cite{pereira2019learning}\cite{clever2017cocomopl}\cite{mainprice2015predicting}\cite{kiumarsi2017optimal}.
	For instance, in \cite{Thuruthel_2017}, the authors applied a NARX neural network to learn the forward dynamics of a soft robot manipulator. Using the learned model, they designed a trajectory optimization method for predictive control of the manipulator. In a subsequent study, the authors improved their controller by implementing a closed-loop scheme using model-based reinforcement learning \cite{thuruthel2018model}.
	Feedforward network models have  improved over the past decade, with increased performance in classification, detection and recognition tasks. One common type of networks is based on Convolutional layers, which works efficiently on the analysis of image data. Conversely, Recurrent Neural Networks (RNN) are appropriate for sequence data, such as video, audio and text, where time is an explicit variable. 
	The so-called \textit{non-linear autoregressive network with exogenous inputs (NARX)} is a type of RNN that can be applied to sequential data such as  the movement dynamics of our robotic eye.
	
\begin{figure}[ht]
		\centering
		\includegraphics[width=0.75\columnwidth]{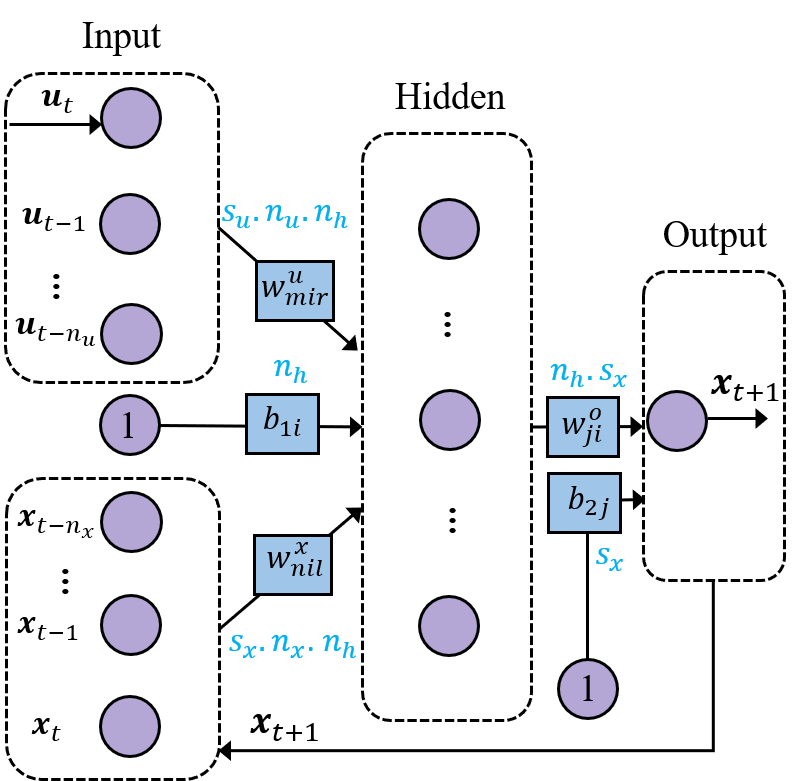}
		\caption{The architecture of the fully-connected NARX model with input, hidden and output layers. Number of hidden units, $n_h$ = 55; input memory: $n_u$ = 1; $n_x$ = 3; degrees of freedom of the state: $s_x$ = 6; degrees of freedom of the input: $s_u = 6$.}
		\label{fig:NARXArchitecture}
	\end{figure}

	The NARX model takes the current and past control signals, $\mathbf{u}_{t-n_u:t}$ (motor commands) and the current and past state, $\mathbf{x}_{t-n_x:t}$ from the nonlinear system as its inputs to predict the state for the next time step (See Fig. \ref{fig:NARXArchitecture}, where  the size of network's memories of the control commands and states is $n_u+1$ and $n_x+1$, respectively (in our model $n_u = 1$ and $n_x = 3$). 
	In discretized time,  the NARX model can thus be written as: 
	\begin{equation}
		\mathbf{x}_{t+1}=f(\mathbf{x}_{t-n_x:t}, \mathbf{u}_{t-n_u:t}).
		\label{eq:NARXdef}               
	\end{equation}
	Fig. \ref{fig:NARXArchitecture} shows the architecture of the applied network in our model. $n_h$ is the number of neurons in the hidden layer. $s_x$ and $s_u$ are the sizes of the input vector $\mathbf{u}$ and state vector $\mathbf{x}$ respectively. $\mathbf{u}_{t-n_u:t}$ and $\mathbf{x}_{t-n_x:t}$ are the inputs to the network. $\mathbf{x}_{t+1}$ is the output of the network. $b_{1,i}$ and $b_{2,j}$ are the bias weights for the hidden layer and output layer, respectively. The weights $w^u_{mir}$ connect the input to the hidden layer, $w^x_{jil}$ connect the current and past states to the hidden layer and $w^o_{ji}$ connect the hidden layer to the output layer  ($i \in [1, n_h]$, $j \in [1, s_x\}$, $m \in [1,s_u]$, $r \in [1,n_u]$, $l \in [1,n_x]$).
	The output of each layer is computed by applying nonlinear (sigmoid) ($f_1$) and linear ($f_2$) activation functions, respectively. The biases and weights of the network  will be tuned during network training using the \textit{Levenberg-Marquardt backpropagation} technique.
	
	The output of neuron $i$ at time $t$ in the hidden layer, $H_i(t)$, is computed by:
	\begin{equation}
		H_i(t)=f_1 \left( \sum_{m=1}^{s_u}\sum_{r=n_u}^{0} w^u_{mir} \mathbf{u}^{(m)}_{t-r}+ \sum_{j=1}^{s_x}\sum_{l=n_x}^{0} w^x_{jil} \mathbf{x}^{(j)}_{t-l} +b_i\right)
		\label{eq:NARXhiddenlayer}               
	\end{equation}
	where the notation $\mathbf{v}^{(k)}$ indicates the $k$th entry of vector $\mathbf{v}$.
	The output of the network is determined by:
	\begin{equation}
		\mathbf{x}^{(j)}_{t+1}=f_2 \left( \sum_{i=0}^{n_h} w^o_{ji} H_i(t) +b_j\right).
		\label{eq:NARXoutputlayer}               
	\end{equation}
	%
	\section{Optimal Control}
	\label{sec:OptimalControl}
	Here, we describe the terms of the cost functions in Eq. (\ref{eq_probdef2}) that were used to optimize the controllers for the \textit{linear} and \textit{nonlinear} model approximations. Saccades should reach the goal as fast and as accurately as possible, while consuming the least amount of '\textit{metabolic}' resources, i.e., energy. Taken together, this leads to the inclusion of three cost functions in the optimal control. 
	%
	\subsubsection*{1. Saccade duration ($J_D$)}
	\textit{Saccade duration} quantifies the time, $T$, 
	it takes the eye to move from the initial to the final state. The longer it takes, the higher the cost, as saccades are supposed to reach the goal in minimum time. The duration cost, $J_D(T)$, is defined by a hyperbolic discount function\cite{Akhil2021}\cite{shadmehr2012}\cite{shadmehr2010temporal}: 
	\begin{equation}
		J_D(T) = 1-\frac{1}{1+\beta T}
	\end{equation}
	%
	%
	\subsubsection*{2. Saccade accuracy ($J_{A}$)}
	Importantly, the eye should reach the goal state $\mathbf{x}_G$ at time $T$ as accurately as possible, and once it gets there, it should stay on the target, i.e., at zero velocity and acceleration if the target is not moving. Thus, the accuracy cost, $J_A$, was defined as the Euclidean norm of the difference between the final 3D eye orientation reached by the controller and the desired goal in 3D, with zero torsion,  $\mathbf{x}_G = (0,G_y,G_z,0,0,0)$. (Note that the primary position is not known a-priori; we therefore expressed the 3D cost in laboratory coordinates).  We also included a penalty for any change in state over a window $W = 5$ samples as soon as the eye reached the goal at time $T$: 
	\begin{equation}
		J_{A}(\mathbf{x}) =\sum_{t=1}^{W}{||\mathbf{x}_G-\mathbf{x}_{T+t}||^2}.
		\label{eq::accuracy}
	\end{equation}
	%
	%
	\subsubsection*{3. Energy consumption ($J_E$)}
	We assume that the total energy consumption by the saccade is proportional to the actuators' angular velocities. As the time steps are uniform, angular velocities can be approximated by differences between angular positions, and the energy cost, $J_E$, can be written as \cite{shadmehr2012}:
	\begin{equation} 
		J_{E}(\mathbf{u})=\sum_{t=1}^T||\mathbf{u}_t-\mathbf{u}_{t-1}||^{2}
		\label{eq:energy}
	\end{equation}

	\subsection{Linear control} 
	%
	Optimal control minimizes a total weighted cost function in order to yield optimal speed-accuracy performance of the system regarding trajectory formation and energy expenditure. The inner optimization loop of Eq. (\ref{eq_probdef2}) is written as:
	\begin{align*}
		& \underset{\mathbf{u}_{0:T}}{\text{min}}\hspace{0.01\textwidth} J(\mathbf{x},\mathbf{u},T) = 
		\sum_{\alpha \in \{D,A,E\}} \lambda_{\alpha} J_{\alpha}(\mathbf{x}, \mathbf{u},T) \\
		%
		& \text{\textbf{subject to}} \notag  \\
		&\hspace{0.1\textwidth} \textbf{x}_{t+1} = \textbf{A}\textbf{x}_t + \textbf{B}\textbf{u}_t \\
		&\hspace{0.1\textwidth} \textbf{y}_t = \textbf{C}\textbf{x}_t, \hspace{0.05\textwidth} t = 0,1,...,T \\
		&\hspace{0.1\textwidth} \mathbf{u}_t \geq 0 
		& \numberthis \label{eq:optmodellinear}
	\end{align*}   
	where $\alpha \in \{D,A,E \}$ indicates the cost functions, and $\lambda_D$, $\lambda_A$ and $\lambda_E$ are the weights associated with cost terms $J_D$, $J_A$ and $J_E$, respectively. In Eq. (\ref{eq:optmodellinear}), \textit{T} is a fixed value. The outer optimization loop of Eq. (\ref{eq_probdef2}) then computes the optimal duration $T^*$ for which the solution of Eq. (\ref{eq:optmodellinear}) is minimal. 

	Because the problem can be formulated in terms of quadratic costs with linear constraints, we applied Matlab's \textit{quadprog} program to solve Eq. (\ref{eq:optmodellinear}). The result from the solver eventually provides the optimal inputs for the nonlinear simulator,  which produces the optimal trajectory for the specified goal. 
	%
	\subsection{Non-linear Control}
	\label{subsec:nonlinearmodel}
	In the non-linear approach we approximated the input motor command $\mathbf{u}_t$ by a Gaussian Mixture Model (GMM)
	
	\begin{equation}
		\mathbf{u}_t =\sum_{i=1}^N \boldsymbol{\mu}_i\varphi^i (t)
		\label{eq:GMM}               
	\end{equation}
	%
	%
	where $\boldsymbol{\varphi}^i(t)$ are Gaussian functions centered at distinct times and  $\boldsymbol{\mu}_i$ are vectors of coefficients (of dimension $6\times1$) to be found using a non-linear constrained optimization solver. More specifically, the Gaussian functions are defined as:
	
	\begin{equation}
		\varphi_i(t)=\frac{\exp\left[\left( \frac{t-c_i}{h}\right)^2 \right] }{\sum_{m=1}^N \exp \left[\left( \frac{t-c_m}{h}\right)^2 \right] }
		\label{eq:GMM2}               
	\end{equation}

	where $c_i$ is the center of the $i^{th}$ Gaussian and $h$ is the common standard deviation.
	Defining the $6 \times N$ matrix $\mathbf{M}= \left[ \boldsymbol{\mu}_1 \cdots \boldsymbol{\mu}_N\right]$, and the $N \times (T+1)$ matrix $\boldsymbol{\Phi} = \left[\boldsymbol{\varphi}^1_{0:T}; \cdots; \boldsymbol{\varphi}^N_{0:T} \right]$, where $\boldsymbol{\varphi}^i_{0:T}$ is the vector with the $i$th Gaussian samples at the discrete times, the control trajectory can be written as
	\begin{equation}
		\mathbf{u}_{0:T} = \mathbf{M}\boldsymbol{\Phi} 
		\label{eq:GMM1}               
	\end{equation}
	
	Thus, instead of finding values for the $6\times (T+1)$ points of a trajectory $\mathbf{u}_{0:T}$, we computed the $6 \times N$ values of the coefficients of the Gaussian Mixture, $N<T$.
	
	The three steps followed in the non-linear approach are as follows: (i) data collection, (ii) computation of the forward dynamics, and (iii) trajectory optimization. 

	With this representation, the inner optimization problem of Eq. (\ref{eq_probdef2}) can be written as

	\begin{align*}
		& \mathbf{M}^{*}=\underset{\mathbf{M}}{\text{argmin}}\hspace{0.01\textwidth} J(\mathbf{x}, \mathbf{u},T) = \sum_{\alpha \in \{D,A,E\}} \lambda_{\alpha} J_{\alpha}(\mathbf{x}, \mathbf{u},T) \\
		& \text{\textbf{subject to:}} \notag  \\
		&\hspace{0.1\textwidth} \mathbf{x}_{t+1}=f(\mathbf{x}_{t},\mathbf{u}_{t}), \quad t = 0,1,...,T\\
		& \hspace{0.1\textwidth}\mathbf{u}_{0:T} =\mathbf{M}\boldsymbol{\Phi}, \quad \mathbf{u}_{t}\geq 0
		\numberthis 
		\label{eq:optmodelnonlinear}
	\end{align*}  
	where each cost term, $J_\alpha$, is multiplied by a fixed weight $\lambda_{\alpha}$ and the next state, $\mathbf{x}_{t+1}$, is computed by the nonlinear NARX model (\ref{eq:NARXdef}). As in the linear optimium control case, $T$ is optimized in the outer loop of Eq. (\ref{eq_probdef2}).
	
	We applied MATLAB's function \textit{fmincon}' with the `\textit{sqp}' solver to optimize Eq. (\ref{eq:optmodelnonlinear}).
	\begin{figure}[ht]
		\centering
		\includegraphics[width=1\columnwidth]{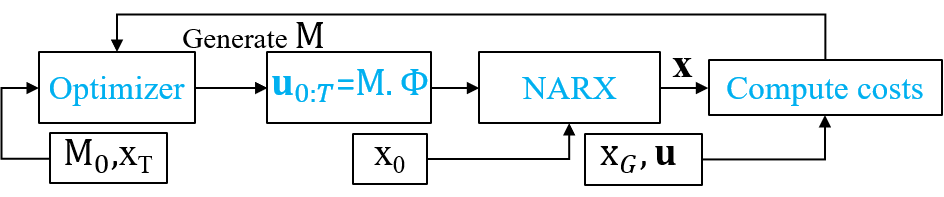}
		\caption{In the nonlinear optimal control approach, the optimizer generates $\mu$ and $\mathbf{u}$ as inputs for the NARX model. From the NARX output, the cost is computed according to the optimization scheme of Eq. \ref{eq:optmodelnonlinear}.}
		\label{fig:optimizationflowchart}
	\end{figure}
	%
	A diagram of the optimization algorithm is shown in Fig. \ref{fig:optimizationflowchart}. We initialized the solver with parameter $\mathbf{M} = 2$ and initial state $\mathbf{x}_0$. The NARX model simulates the behavior of the nonlinear eye system given the input trajectory $\mathbf{u}$ and eye orientation $\mathbf{x}$, and generates the corresponding next states. Next, 
	the solver computes the cost of the resulting saccade. The loop continues for a fixed number of iterations, or until a stop condition is met.
	
	%
	One of the challenges for the optimization is to set appropriate values for the cost weights $\lambda_\alpha$. We first estimated the range of candidate values, after which by trial and error we manually found appropriate values for the $\lambda$'s .
	
	The full optimization procedure, including the inner and outer optimization loops of Eq. (\ref{eq_probdef2}), is shown in Alg.~\ref{Alg1:opt}. It takes the initial and desired orientation of the eye ($\mathbf{x}_0, \mathbf{x}_G$) as inputs. Then, for each possible saccade duration $0 < T \leq T_{\max}$, it computes the optimal motor controls and the cost of that solution. Finally, it chooses the solution with the lowest cost. 
	The duration cost $J_D$ penalizes the saccade duration and is computed by dividing the saccade duration into $d$ portions, and run the optimizer for different durations ($T$) over the range $T=[ts,2ts,3ts,\cdots,T_{max}]$ for a given saccade, where $ts=T_{max}/d$. In this way, we made $d$ runs for every saccade (see lines 2 to 6 in Alg. \ref{Alg1:opt})  instead of searching for all possible values; for computational efficiency.
	$c$ is used to specify the number of basis functions in the GMM, and the three $\lambda_\alpha$'s are the cost weightings.

	%
	\begin{algorithm}[ht]
		\caption{ $Trajectory\_Optimization()$ }
		\label{Alg1:opt}
		\SetAlgoVlined
		\DontPrintSemicolon
		\KwIn{$ \mathbf{x}_0, ~ \mathbf{x}_{G}  $ \color{blue} Initial and final orientation \color{black} }
		\nl $param \gets \lambda, ~d$ {\color{blue} Initialize $param$.} \;
		\For{$i <$  d}
		{	
			\nl $Basis\_Num \gets c*T_i$ {\color{blue} Basis length for GMM.} \;
			\nl $\boldsymbol{\Phi} \gets GMM(Basis\_Num)$ {\color{blue} Create $\boldsymbol{\Phi}$ from GMM.} \;			
			\nl $ \mathbf{M}_0 \gets CreateMu(Basis\_Num)$ {\color{blue} Initialize $\mathbf{M}$.} \;		
			\nl $ cost_i \gets Solver(policy\_cost(\mathbf{M}_0,\boldsymbol{\Phi},\lambda)$ {\color{blue} Call optimizer with policy cost.} \;  		    					
		}
		\nl $ \mathbf{M}^* \gets min(cost)$ {\color{blue} Find optimum $\mathbf{M}$ given the costs.}\;			
		\nl $ \mathbf{u}^* \gets \mathbf{M}^* \cdot \boldsymbol{\Phi} $ {\color{blue} Optimal trajectory.} \;			
		\end {algorithm}
		%
		In Alg. \ref{Alg2:Policy_cost} the function $Policy\_cost$ is used in our optimizer with $\mathbf{M}$, $\lambda$ and $\boldsymbol{\Phi}$ as its inputs and total cost $J$ as the output. After simulating the eye behavior using the motor commands in the NARX model, we compute the asociated costs in line 3. The total cost is subsequently computed as the weighted sum of the costs (line 4).
		
		In our experiments, the following values were set: $d=10$, $maxiter=15$ (the maximum number of iterations for the solver) and $c=1/1.4$. 
		We further constrained the maximum saccade duration to $T_{max}= 210 ms$. It should be noticed that although we use discretized duration values in our optimization process, to be more realistic, we compute the real duration to represent our simulation result. To do so, we consider the time when the velocity of the saccade goes down and reaches to 0.1 of the peak velocity.
		%
		\begin{algorithm}[ht]
			\caption{ $Policy\_cost(\mathbf{M},\boldsymbol{\Phi},\lambda)$ }
			\label{Alg2:Policy_cost}
			\SetAlgoVlined
			\DontPrintSemicolon
			\KwIn{$ \mathbf{M},\boldsymbol{\Phi},\lambda$ \color{blue} Inputs  }
			\KwOut{ $J$ \color{blue}  Total cost}		
			\nl $\mathbf{u} \gets \mathbf{M}*\boldsymbol{\Phi}  $ {\color{blue} Create trajectory.} \;	
			\nl $\mathbf{x} \gets NARX(\mathbf{u})$ {\color{blue} Simulate NARX model.} \;
			\nl $J_\alpha \gets compute\_cost(\mathbf{x})$ {\color{blue} Sub-costs} \;
			\nl $J \gets  \lambda_{\alpha}.J_{\alpha}$ {\color{blue} Total cost} \;
			\end {algorithm}
			\subsection{Controlling pretension}
			\label{subsec::pretension}
			
			Because of the redundancy in controlling the 3D orientation of the eye with 6 motors, the same  orientation can be achieved by (infinitely many) different motor-angle combinations, as the amount of co-contraction of antagonistically acting muscles is undetermined.
			Therefore, an important feature to control is the amount of pre-tension (the set of initial motor angles), such that the eye is in equilibrium at all orientations, and is able to reach any orientation in the oculomotor range at optimal speed and minimal effort without the cables going slack during eye movements. We thus optimized the initial motor angles ($\mathbf{u}$) in the following way: 
			\begin{align*}
				& \underset{\mathbf{u}}{\text{min}}\hspace{0.01\textwidth} \lVert \mathbf{u} \rVert^2 \\
				& \text{\textbf{subject to}} \notag  \\
				&\hspace{0.1\textwidth} \textbf{f}(\mathbf{x},\mathbf{u}) > 0 \\
				&\hspace{0.1\textwidth} \mathbf{\tau}_{k}(\mathbf{x},\mathbf{u}) = 0 \\
				&\hspace{0.1\textwidth} \mathbf{u}_{agonist} + \mathbf{u}_{antagonist}> 2\theta
				& \numberthis \label{eq:optequilibrium}
			\end{align*}
			where $\mathbf{f}$ if the tension at each cable, $\boldsymbol{\tau}_k$ is the total torque resulting from the cable tensions, and $\theta$ is a minimum bound for the average motor angle for the three antagonistic motors pairs ($\mathbf{u}_{agonist}$ and $\mathbf{u}_{antagonist}$). These pairs are MR-LR, IR-SR and IO-SO. The value for $\theta$ was chosen such that the cables would not go slack throughout the trajectory. Through trial and error we found that values around 2 radians worked well (e.g., dashed lines in Fig. 15). This value depends on the radius of the spindle around which the cables are wound. The other constraints of the optimization were selected to ensure positive forces in the cables at the equilibrium state $\mathbf{x}$, while the total elastic torque on the system is zero (see Eqns.~\ref{full eye torque} and \ref{eq:force} ).
			It should be noted that in the linear control method for the continuous saccade-set simulations the pretension had to be tuned for the new starting point after every saccade (see Sec. \ref{sec:linearizedmodel}). In contrast, for the non-linear control this was only needed for the first starting point; in the zero-initial simulations, all saccades started from straight-ahead, so we needed to set the pre-tension only once for all saccades. 
			
			\section{Simulation Results}
			\label{sec:simulationresult}
			To evaluate the linear and nonlinear approaches, we will analyse and compare several output parameters of the system regarding the 3D kinematic (Listing's plane) and dynamic (velocity profile) behavior. 
			%
			\subsection{Simulation Setup}
\begin{figure}[t]
				\centering
				\subfigure{\includegraphics[width=0.7\columnwidth]{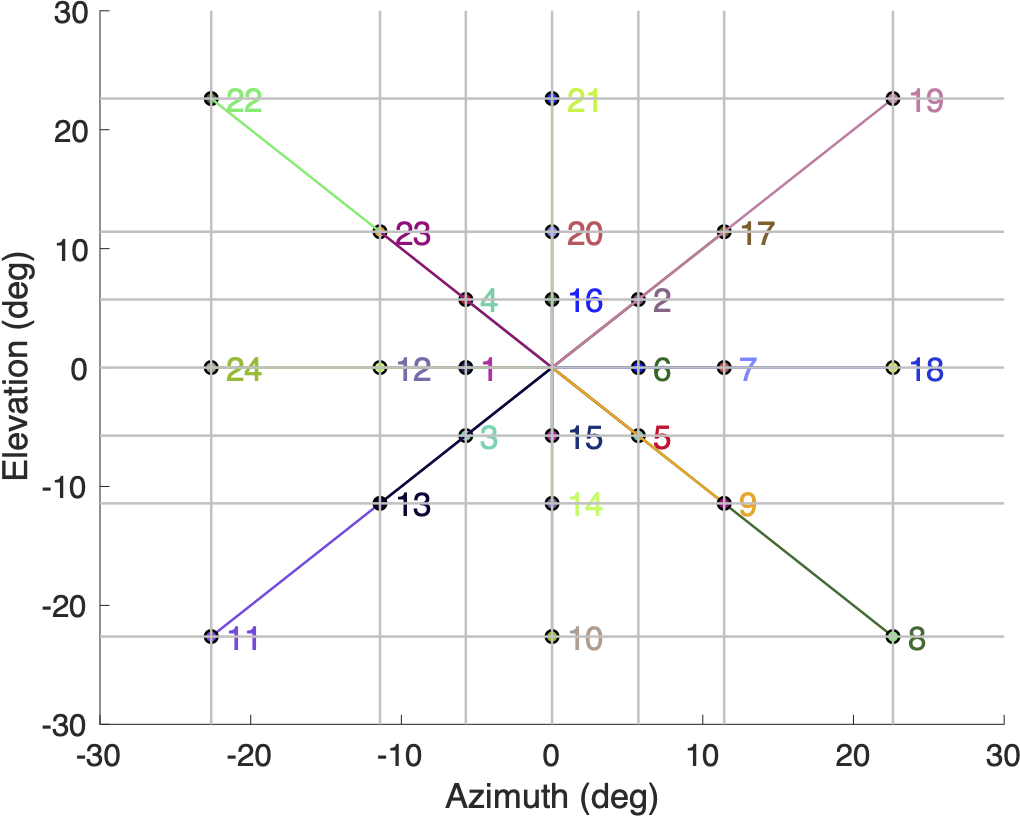}}\\
				\subfigure{\includegraphics[width=0.7\columnwidth]{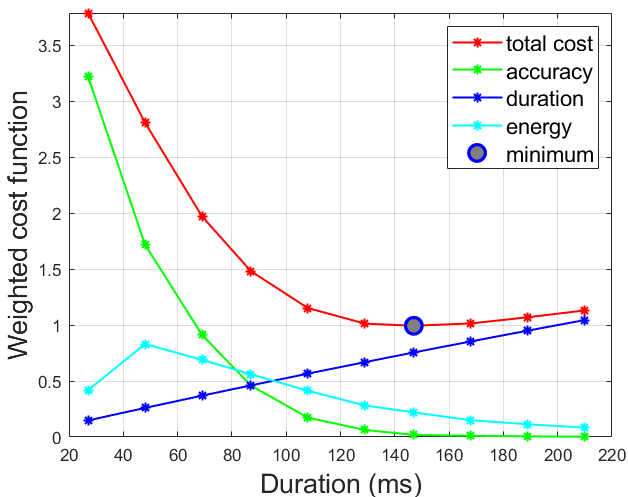}}
				\caption{\textbf{Top}: The 24 oblique goal directions as azimuth (rightward positive) and elevation (upward positive) angles in the amplitude range from 5 to 30 degs, as used for the zero-initial and continuous saccade tests in our experiments. 
					\textbf{Bottom}: Cost functions for a rightward horizontal saccade of 22$^o$ starting from the origin in the trajectory optimization procedure. The large blue dot indicates the minimum total cost of the trajectory, and is found at T = 147 ms.} 
				\label{fig:sacc_cost_motor}
			\end{figure}

			To enable a fair comparison between the two methods, we created a saccade set containing 24 different target locations in horizontal, vertical and oblique directions with amplitudes between 5 and 33 degrees from straight-ahead (see Fig. \ref{fig:sacc_cost_motor}a).  
			We performed two different saccade sequences: a \textit{zero-initial} sequence, where every saccade started from the origin ($[0,0,0]$), and a \textit{continuous} sequence, where the next saccade started from the final orientation of the previous saccade, and so on.
			All simulations were performed in MATLAB 2021 on a laptop with the Windows 10 operating system, and 16GB Ram and a CPU core i7.

			The three cost functions, $J_\alpha$, for our optimization equations (Eq. \ref{eq:optmodellinear} and Eq. \ref{eq:optmodelnonlinear}) were identical for the linear and nonlinear methods, but their weights $\lambda _\alpha$ were set at different values due to differences in time sampling, forward model structure, and optimization algorithms. The cost multipliers were manually calibrated with the aim to achieve human-like dynamic characteristics for the eye system. 
			The selected values for the three multipliers are given in Table \ref{Tbl:parameters}.
			An example of the behavior of the three cost functions for a 22 deg horizontal saccade, together with the total cost (red), executed at ten different saccade durations between 30 and 210 ms, is shown in Fig. \ref{fig:sacc_cost_motor}b). The optimal saccade duration, T is found at the minimum of the convex curve representing the total cost, i.e., at $T$=147 ms. 
			
			\begin{table}[ht]
				\centering
				\caption{Cost weightings for both controllers.}
				\label{Tbl:parameters}
				\begin{tabular}{rcccl}
					Weight & $\lambda_A$ & $\lambda_D$ & $\lambda_E$ \\ 
					\hline
					Linear    & 0.33 &1.00 & 0.67\\ 
					\hline
					Nonlinear & 1.00 & 0.04 & 0.002\\ 
				\end{tabular}
			\end{table}
			%
			%
			\begin{figure}[t]
				\centering
				\includegraphics[width=0.95\columnwidth]{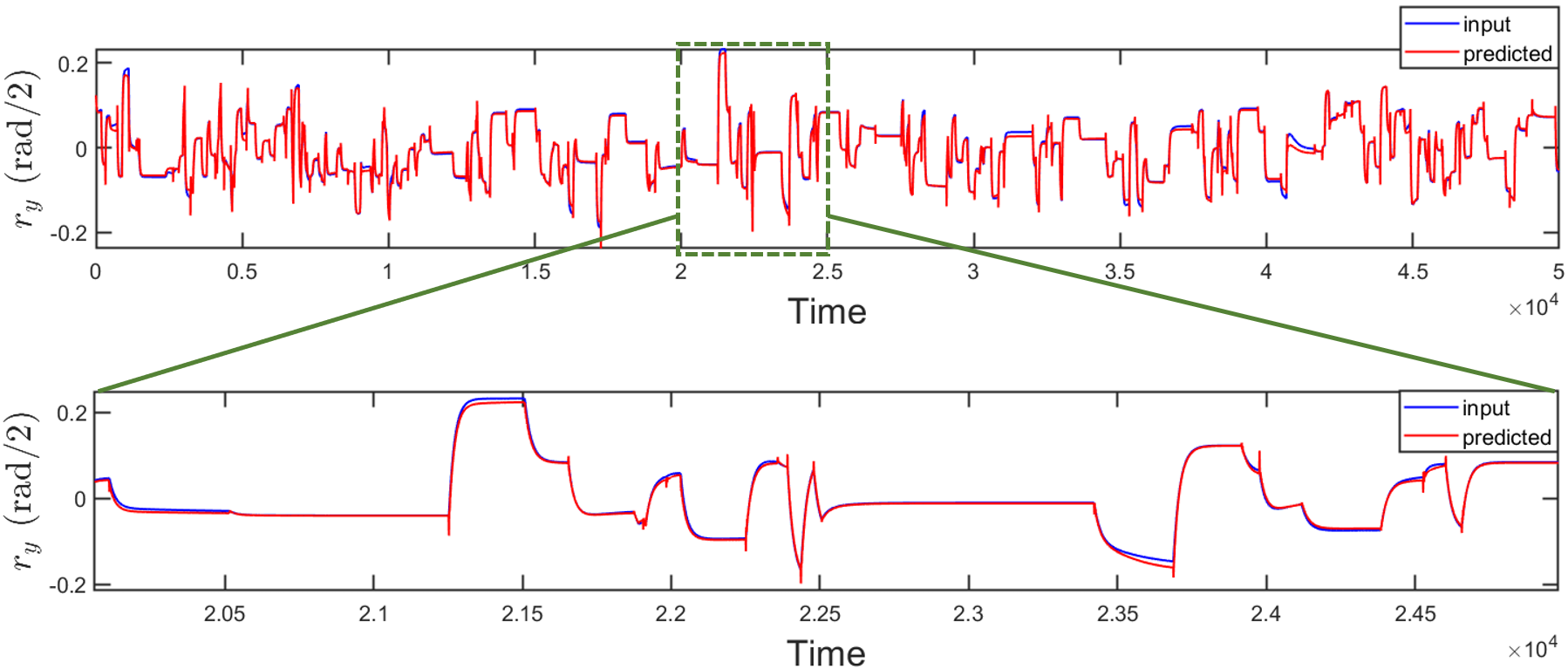}
				\caption{Eye orientation along the $y$-axis for a sample input signal (blue trace) of 50,000 samples (150,000 ms), and the predicted signal (red) for the non-linear trained NARX model. Inset: same data on an expanded scale. RMSE on the test set of $0.06$ and $R^2=0.97$ between data and prediction are indicative for an excellent approximation.}
				\label{fig:trainedRNNtest}
			\end{figure}
			%
			\subsection{Non-linear Model Learning}
			
			\begin{figure*}[ht]
				\centering
				\subfigure[]{\includegraphics[width=0.65\columnwidth]{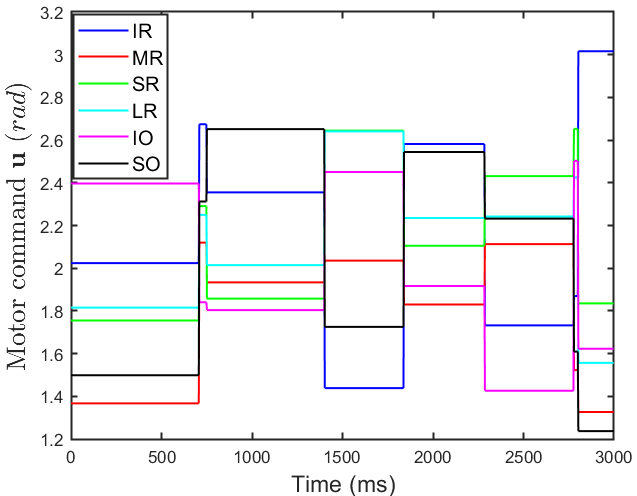}}~~
				\subfigure[]{}{\includegraphics[width=0.65\columnwidth]{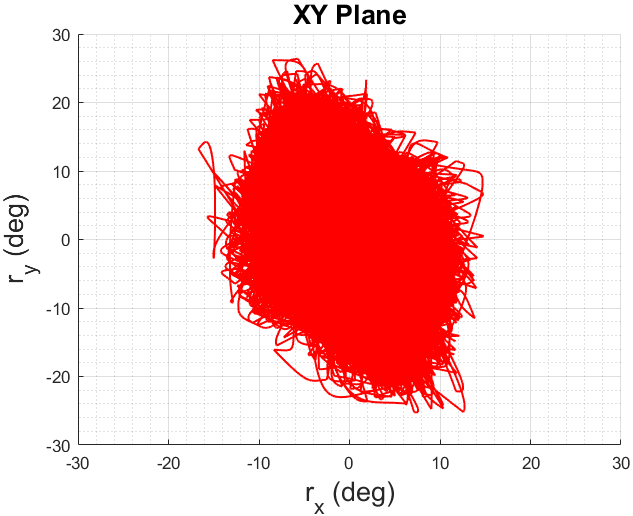}}~~
				\subfigure[]{\includegraphics[width=0.65\columnwidth]{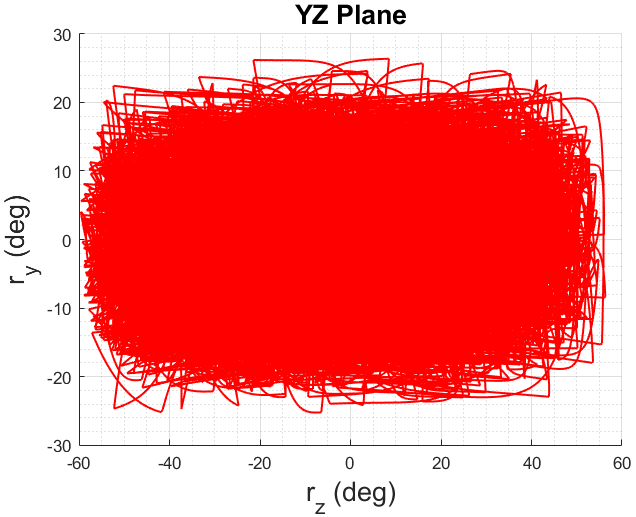}}
				\caption{\textbf{Left}: An illustrative selected section of random motor responses over 3000 ms (out of $2\times10^6$ ms), which were used as the training set for the NARX network. Motor commands are in radians, within a range [0-3]. \textbf{Center and right}: The corresponding eye orientations, represented in degrees by their rotation-vector components in the $(xy)$ and $(yz)$ planes of the laboratory frame (a sub-sample of the total data set). Note that the vertical-torsional range of the system is smaller than the horizontal range, and that the data are slightly tilted in the $xy$ plane.}
				\label{fig:datasetdistribution}
			\end{figure*}
			
			To train the recurrent neural network we used a selected dataset of saccadic movements generated by the nonlinear simulator. To create the dataset we produced a continuous sequence of random saccades at 1 ms sampling rate (see Sec. \ref{sec:cabledrivensye}). The total dataset length is $2\times 10^6$ ms, which, as can be seen in Fig. \ref{fig:datasetdistribution}, covers a wide range of the workspace. For computational reasons, we reduced the size of this dataset by down-sampling the signals to 3 ms time-intervals.
			
			By feeding the data set (see Sec. \ref{subsec:nonlinearmodel}) to the NARX model, the best result was achieved after 96 epochs with MSE after training = 0.0018 $(rad/2)^2$.
			Figure \ref{fig:trainedRNNtest} illustrates the result (for a 3000 ms sequence) of testing the prediction of the trained NARX network (red trace) on a random set of saccades (blue trace), which verifies how well the network learned the forward dynamics of our nonlinear robotic eye model. 
			
			%
			\subsection{Kinematic behavior Analysis}
\begin{figure}[t]
				\setlength\tabcolsep{1pt}
				\settowidth\rotheadsize{Radcliffe Cam}
				\setkeys{Gin}{width=\hsize}
				\begin{tabularx}{1\linewidth}{l| XXX }
					\rothead{\centering	\textcolor{blue}{Linear}}       
					&   \includegraphics[valign=m]{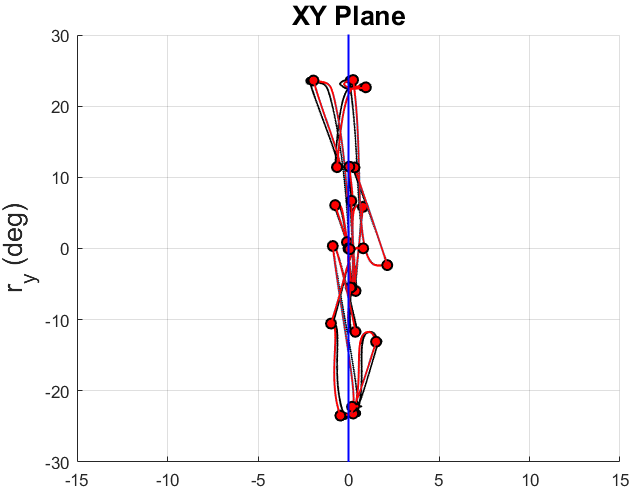}
					&   \includegraphics[valign=m]{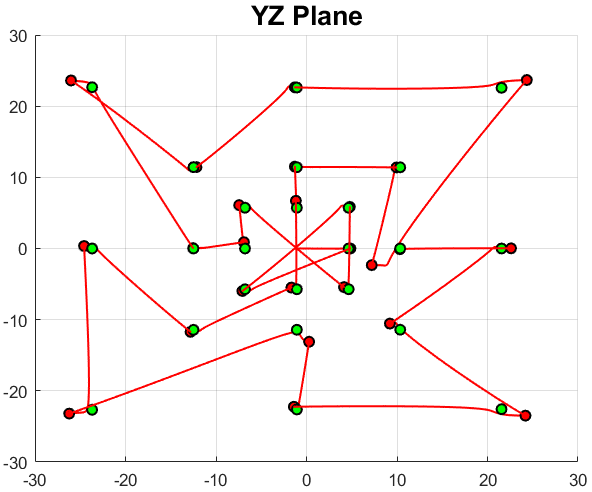}     \\  
					\addlinespace[2pt]
					\rothead{\centering \textcolor{blue}{Nonlinear}} 
					&   \includegraphics[valign=m]{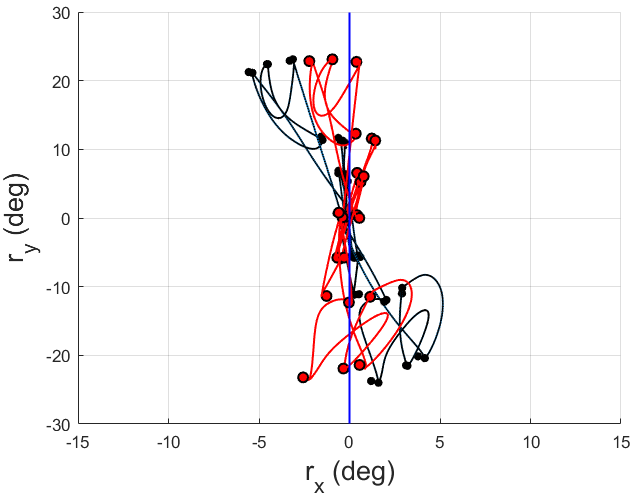}
					&   \includegraphics[valign=m]{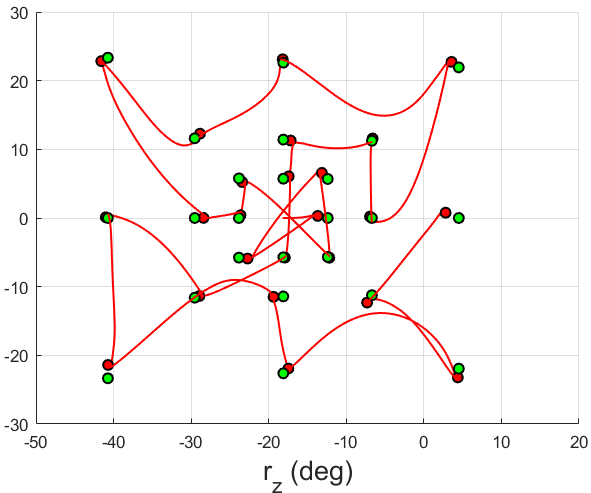}
				\end{tabularx}
				\caption{$xy$ (left) and $yz$ (right) projections of the instantaneous 3D eye-movement trajectories in Listing's coordinates (red) from the continuous saccade sequence. Black: $xy$ data in laboratory frame before rotation about the $z$-axis into Listing's coordinates. \textbf{Top}: linearized approximation, rotation: 2.15 deg; \textbf{Bottom}: nonlinear approach, rotation: 18.11 deg. The blue vertical line in the $xy$ plots indicates Listing's plane; green dots correspond to the actual goals of the saccade set in Listing's frame.}
				\label{fig:xyplanes}
			\end{figure}

			%
			To study the eye-movement kinematics resulting from our controllers, we analyzed the amount of cyclotorsion of the 3D trajectories with respect to Listing's plane. The results are shown in the left-hand panels of Fig. \ref{fig:xyplanes} as a projection of the trajectories of both saccade sets onto the $xy$ plane. Listing's plane is indicated by the blue vertical line. The standard deviations of the cyclotorsional components of the two models, for different movement-amplitude ranges, are presented in Table~\ref{Tbl:comparison}.
			
			Although the differences are small, the nonlinear model yields slightly smaller deviations from LP than the linearized model. 
			\begin{table}[t]
				\centering
				\caption{Standard and maximum deviation of the Listing's plane.}
				\label{Tbl:comparison}
				\begin{tabular}{p{1.5cm}|p{1.5cm}|p{1.6cm}|p{1.6cm}}
					\textbf{Saccade/ Control}              & \textbf{Amplitude}$^{\circ}$       &\textbf{Linear} \qquad    (STD, Max)$^{\circ}$ &\textbf{Nonlinear} (STD, Max)$^{\circ}$ \\ 
					\hline
					\multirow{3}{*}{Zero-initial}  & $<$ 10 &  0.10, 0.19  & 0.12, 0.3   \\ \cline{2-4} 
					& $<$ 20 \& $\geq$10 &  0.41, 0.93  & 0.24, 1.2   \\ \cline{2-4} 
					& $<$ 35 \& $\geq$20 &  1.64, 3.49  & 0.18, 0.69    \\ \hline
					\multicolumn{1}{c|}{\multirow{3}{*}{Continuous}} & $<$ 10            &  0.16, 0.64  & 0.32, 0.96     \\ \cline{2-4} 
					\multicolumn{1}{c|}{}                           & $<$ 20 \& $\geq$10 &  0.59, 1.94  & 0.58, 1.71    \\ \cline{2-4} 
					\multicolumn{1}{c|}{}                           & $<$ 35 \& $\geq$20 &  0.82, 1.71  & 0.67, 1.3       
				\end{tabular}
			\end{table}
			The right-hand panels of Fig.~\ref{fig:xyplanes} show the trajectories of the eye movements from the continuous set in the $YZ$ plane for the two approximations. Note that the oblique trajectories are quite straight, especially for the linearized model data. The trajectories from the nonlinear model are curved at the larger eccentricities. Yet, as shown below, the amount of cross-coupling between the horizontal and vertical eye-movement components is quite significant for the two approaches.  

\begin{figure}
				\setlength\tabcolsep{1pt}
				\settowidth\rotheadsize{Radcliffe Cam}
				\setkeys{Gin}{width=\hsize}
				\begin{tabularx}{1\linewidth}{l| XXX }
					\rothead{\centering	\textcolor{blue}{Linear}}       
					&   \includegraphics[valign=m]{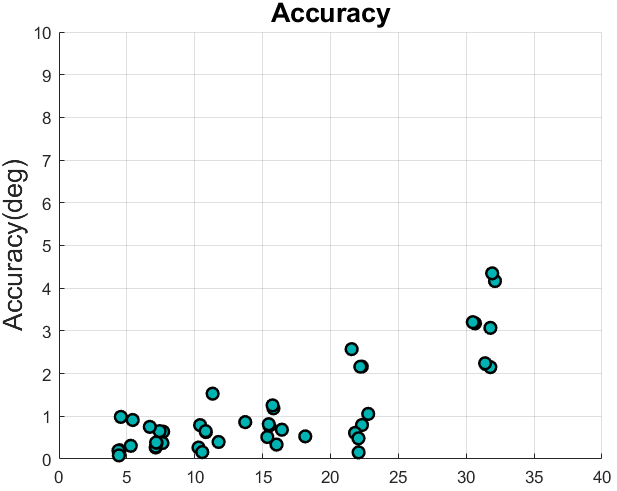}
					&   \includegraphics[valign=m]{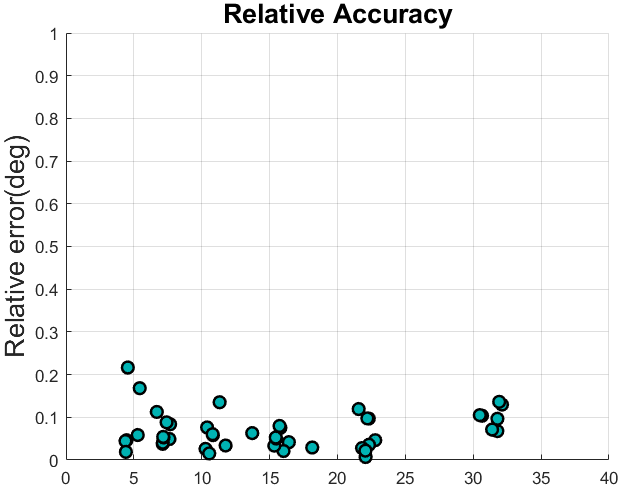}     \\  
					\addlinespace[2pt]
					\rothead{\centering \textcolor{blue}{Nonlinear}} 
					&   \includegraphics[valign=m]{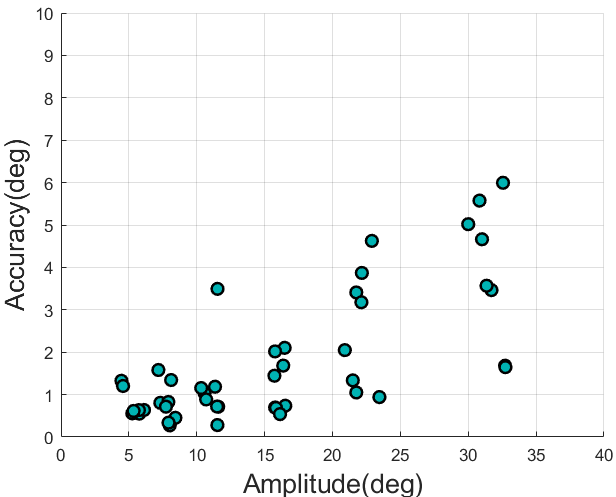}
					&   \includegraphics[valign=m]{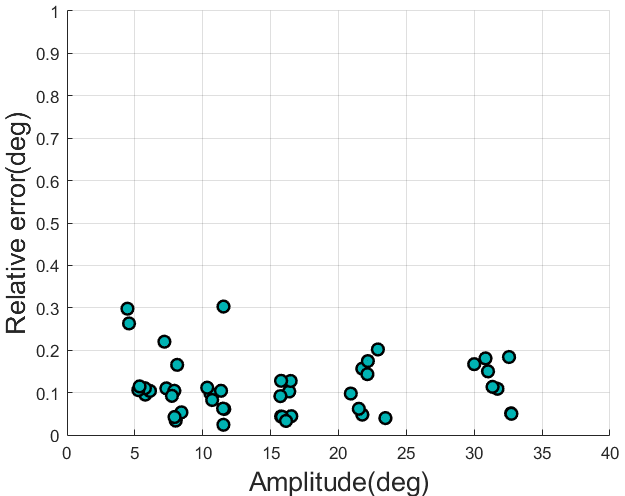}
				\end{tabularx}
				\caption{Absolute error (left) and relative error (absolute error/amplitude; right) of the pooled saccade sets for the linearized (top) and nonlinear controls (bottom). The relative error is about 10\% for the linearized model, and is slightly higher and more variable for the nonlinear model.}
				\label{fig:accuracy}
			\end{figure}

			The accuracy (absolute error) and relative error for the two model approximations are shown in Fig. \ref{fig:accuracy}. The absolute localization error increases nearly linearly with movement amplitude from 0 to about 5 deg. The relative error (absolute error normalized for movement amplitude) hovers around 10\% for both approximations, independent of movement amplitude. This result corresponds to the known inaccuracy of human saccadic eye-movements, which tend to undershoot the target by about 10\% \cite{robinson2022}. In general, the nonlinear control led to slightly larger and more variable relative errors than the linearized model, as shown in table \ref{Tbl:meanSTDRelerr_Velrat}. Both zero-initial and continuous saccade types are considered in the computation. 
			
			\begin{table}[ht]
				\centering
				\caption{Mean and STD for relative gaze error and trajectory straightness.}
				\label{Tbl:meanSTDRelerr_Velrat}
				\begin{tabular}{p{2.0cm}|p{1.7cm}|p{1.7cm}}
					\textbf{}   &\textbf{Linear} \qquad (Mean, STD) &\textbf{Nonlinear} (Mean, STD) \\ 
					\hline
					Relative error ($^{\circ}$)  &  0.08, 0.07  & 0.12, 0.07    \\ \hline
					Straightness ($^{\circ}$) & 0.98, 0.025 & 0.81, 0.22        
				\end{tabular}
			\end{table}
			%
			%
			\subsection{Dynamic behavior Analysis}
			
			\begin{figure}
				\setlength\tabcolsep{1pt}
				\settowidth\rotheadsize{Radcliffe Cam}
				\setkeys{Gin}{width=\hsize}
				\begin{tabularx}{1\linewidth}{l| XXX }
					\rothead{\centering	\textcolor{blue}{Linear}}       
					&   \includegraphics[valign=m]{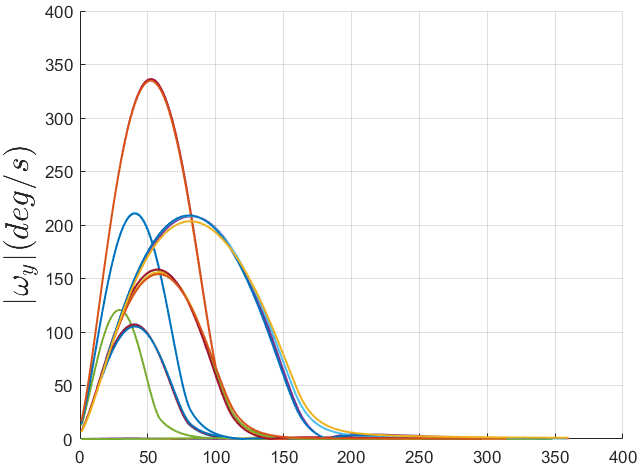}
					&   \includegraphics[valign=m]{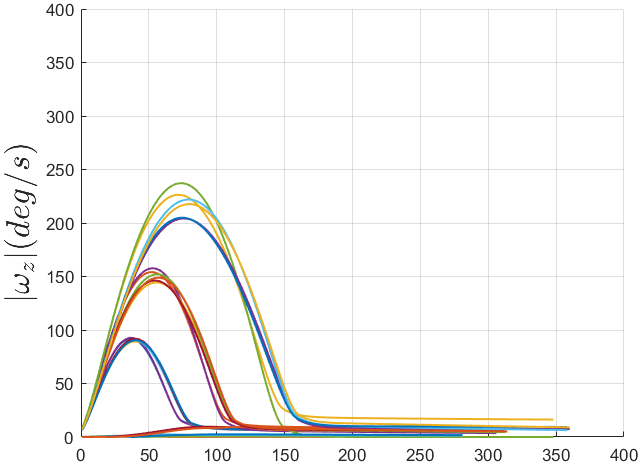}\\  
					\addlinespace[2pt]
					\rothead{\centering \textcolor{blue}{Nonlinear}} 
					&   \includegraphics[valign=m]{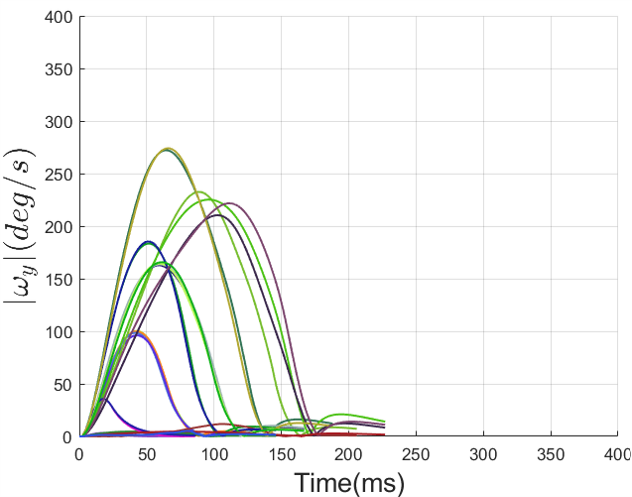}
					&   \includegraphics[valign=m]{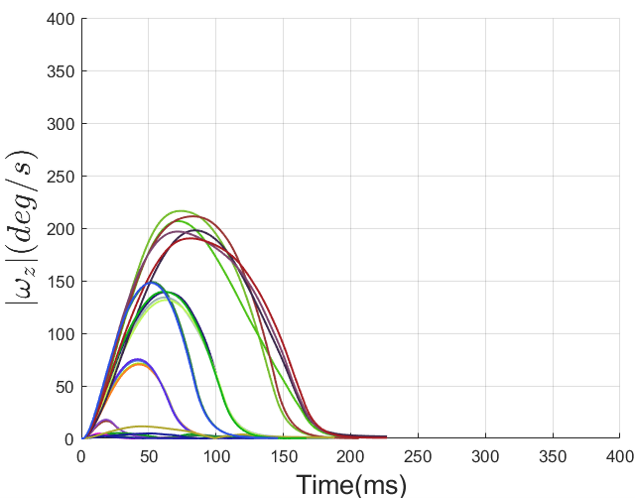}
				\end{tabularx}
				\caption{Velocity profiles of the zero-initial saccade set in the vertical ($y$; left) and horizontal ($z$, right) directions, as created by the linear (top) and nonlinear (bottom) controllers. Line colors correspond to the numbers in the top panel of Fig.~\ref{fig:sacc_cost_motor}.}
				\label{fig:velocityprofile}
			\end{figure}

			\begin{figure}
				\setlength\tabcolsep{1pt}
				\settowidth\rotheadsize{Radcliffe Cam}
				\setkeys{Gin}{width=\hsize}
				\begin{tabularx}{1\linewidth}{l| XXX }
					\rothead{\centering	\textcolor{blue}{Linear}}       
					&   \includegraphics[valign=m]{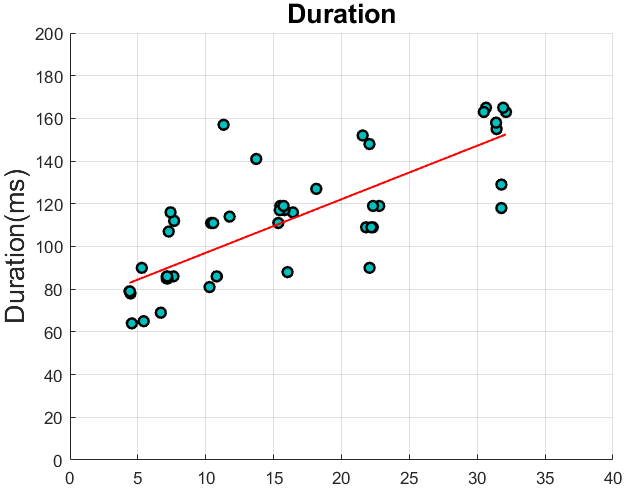}
					&   \includegraphics[valign=m]{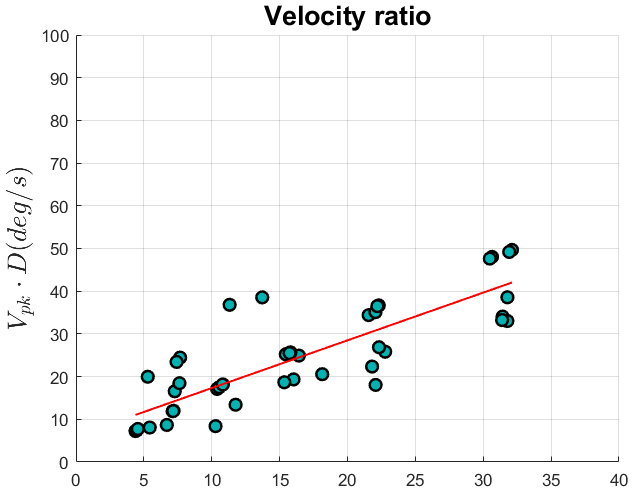}     \\  
					\addlinespace[2pt]
					\rothead{\centering \textcolor{blue}{Nonlinear}} 
					&   \includegraphics[valign=m]{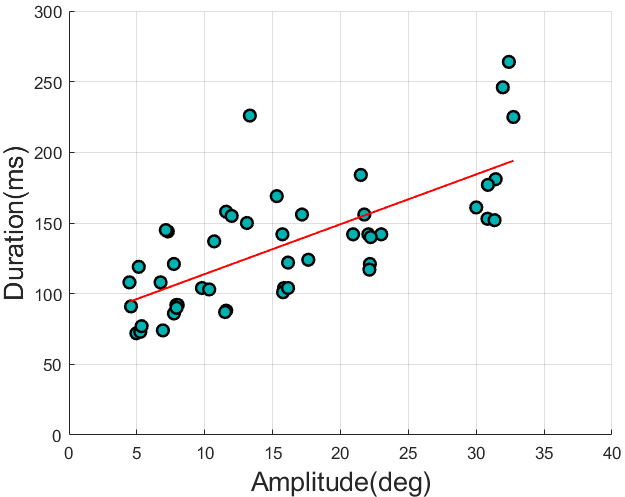}
					&   \includegraphics[valign=m]{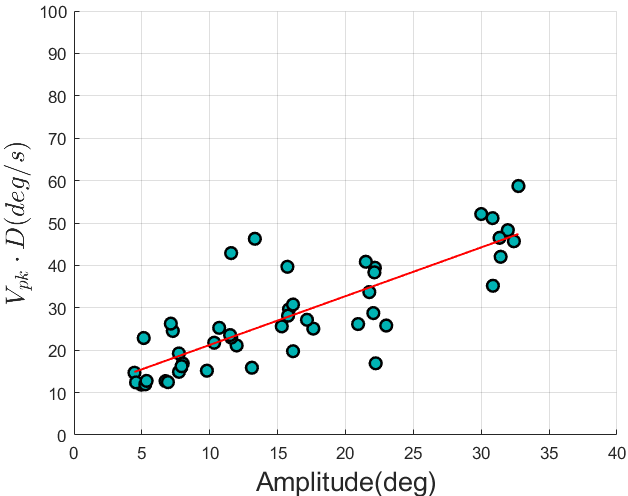}
				\end{tabularx}
				\caption{Main-sequence properties. Left: amplitude - duration relationship. Right: amplitude vs. $V_{pk}\cdot D$ (in deg) for the linear (top) and nonlinear (bottom) controllers. Data include both saccade sets.}
				\label{fig:dur_amp}
			\end{figure}

			Figure~\ref{fig:velocityprofile} shows the velocity profiles for the zero-initial saccade set in the vertical (left) and horizontal (right) directions for the linear (top) and nonlinear (bottom) model approximations. It can be seen that the variability in the velocity profiles of the linearized approach is smaller than for the nonlinear controller, which is in line with the slighlty stronger curvatures in the saccade trajectories for the latter. Note also that the highest peak velocities for the nonlinear control are lower than for the linearized control. Yet, the overall shapes of the velocity profiles are quite similar for the two approaches. Also, the correlations between the horizontal and vertical velocity profiles for the oblique saccades in the set are high, indicating that these components show a considerable amount of cross-coupling (See table \ref{Tbl:meanSTDRelerr_Velrat}), although the correlation is higher for the linear than for the nonlinear control. 
			
			In Fig.~\ref{fig:dur_amp} we show the main-sequence properties of the saccades, pooled for both target sets, for the two model approximations. For both approximations, the movement duration increases with amplitude (left-hand column). As explained above, when velocity profiles are single-peaked (e.g., Fig.~\ref{fig:velocityprofile}), amplitude and $V_{pk}\cdot D$  are expected to be linearly related, as shown in the right-hand column. The variability in the dependent variables is larger for the nonlinear controller than for the linearized controller. Part of the variability is due to the direction-dependence of the saccade dynamics on saccade direction, as they invoke different muscle synergies that influence the movement speed. How these synergies are formed is described next. 
			%
			\subsection{Analysing Muscle Forces}
			
\begin{figure}
				\setlength\tabcolsep{1pt}
				\settowidth\rotheadsize{Radcliffe Cam}
				\setkeys{Gin}{width=\hsize}
				\begin{tabularx}{1\linewidth}{l XXX }
					\rothead{\centering	\textcolor{blue}{Linear}}       
					&   \includegraphics[valign=m]{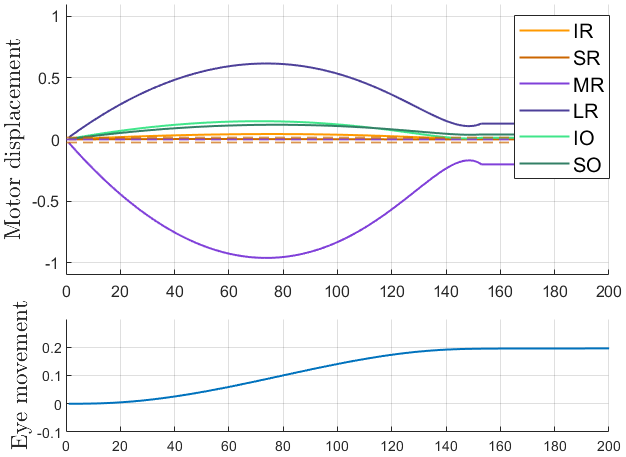}
					&   \includegraphics[valign=m]{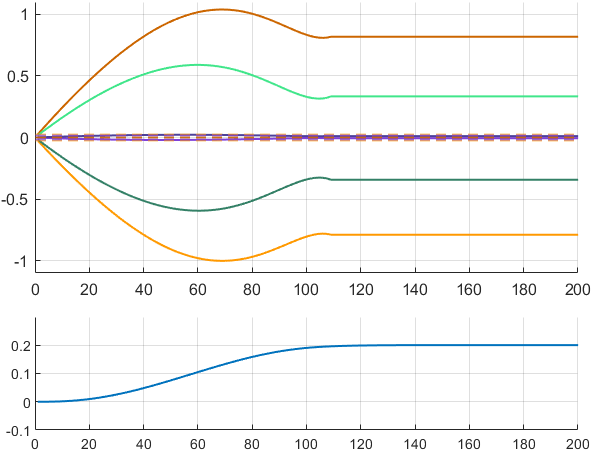}     \\  
					\addlinespace[2pt]
					\rothead{\centering \textcolor{blue}{Nonlinear}} 
					&   \includegraphics[valign=m]{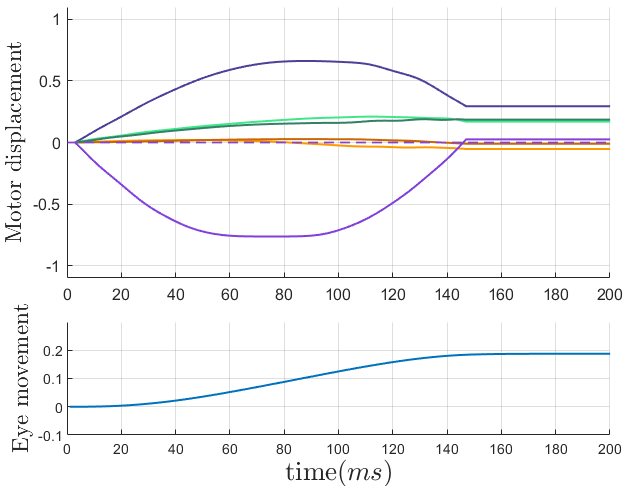}
					&   \includegraphics[valign=m]{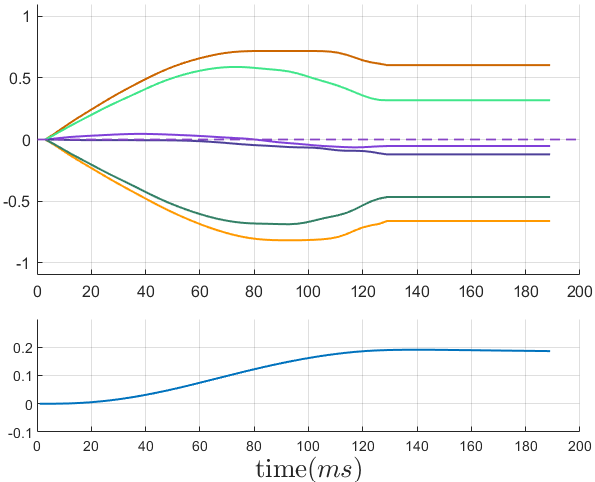}
				\end{tabularx}
				\caption{Changes in motor-command angles (radians) with respect to the pretension values for each tendon, associated with the optimal trajectories for two purely 22.6 deg rightward horizontal (left) and upward vertical (right) saccades, starting from the straight-ahead position with the linear and nonlinear controllers.
					For illustrative purpose, initial motors angles are aligned with zero, but the real values are $[2.01, 1.96, 1.98, 2.03, 1.97, 2.02]$ rad, for all saccades. Panels below each plot show the associated absolute eye-movement trace (in rad/2). Note the different synergies that organize the muscles into agonists and antagonists and that pulse/antipulse durations match the saccade duration.}
				\label{fig:motoranglesL}
			\end{figure}

\begin{figure}
				\setlength\tabcolsep{1pt}
				\settowidth\rotheadsize{Radcliffe Cam}
				\setkeys{Gin}{width=\hsize}
				\begin{tabularx}{1\linewidth}{XX}
					\centering\textcolor{blue}{Linear} &  \textcolor{blue}{\qquad \qquad \quad Nonlinear}\\
					\includegraphics[valign=m]{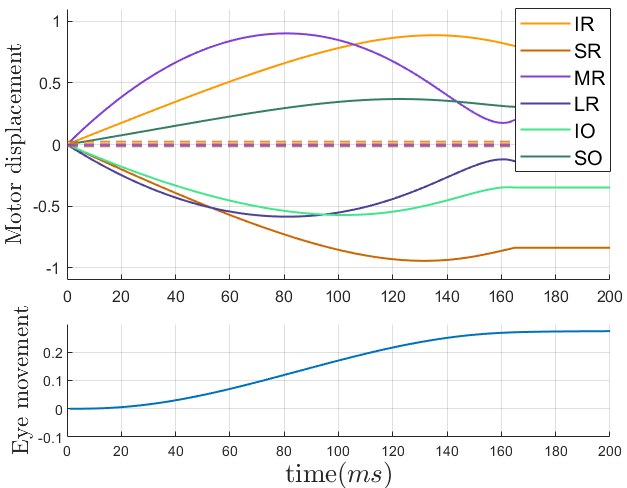} &  \includegraphics[valign=m]{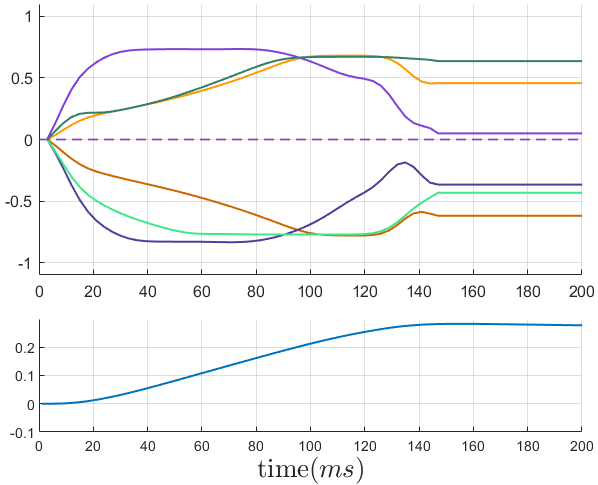}\\
				\end{tabularx}
				\caption{Changes in motor-command angles re. pretension for each tendon associated with the optimal trajectory of a right- and upward oblique saccade to [22.6, 22.6] deg from straight-ahead with the linear and nonlinear controllers. Same format as Fig.~\ref{fig:motoranglesL}. Note that all six muscles are involved in the oblique saccade and that the pulse-step behaviors of the antagonistic muscle pairs are similar for the two methods. Note also the different control dynamics for the horizontal vs. the vertical/torsional system.}
				\label{fig:motoranglesNL}
			\end{figure}

			Figures \ref{fig:motoranglesL} (horizontal and vertical) and \ref{fig:motoranglesNL} (oblique eye movement to the right and upward) show the  motor-control angles as function of time for all six tendons, for the linear and nonlinear model approximations for saccades starting from the straight-ahead direction.  
			Several interesting observations can be made in these figures that apply to both models: 
			First, purely horizontal saccades only involve the activation of the m$_{MR}$ and m$_{LR}$ tendons, while the purely vertical saccades involve the joint action of m$_{SR}$, m$_{IR}$ and m$_{SO}$ and m$_{IO}$ tendons. For oblique saccades, all six tendons are activated.
			Second, the motors appear to clearly act in an antagonistic fashion for all saccade types, as the main activation controls for the involved muscles are in opposite directions \cite{robinson75}\cite{hepp85}\cite{suzuki99}.  
			Third, the control signals to the eye plant from the involved tendons can be characterized as a {\it pulse-step} activation for the agonists, and an anti-pulse/negative step for the antagonists \cite{robinson2022}.
			Fourth, the positive and negative pulses of the activated agonists and antagonists exactly match the saccade durations, whereas the agonist and antagonist post-saccade steps attain increased and decreased values relative to the initial equilibrium pretensions, respectively. 
			Finally, the pulse (and anti-pulse) amplitudes increase with the saccade amplitude.
			Thus, the net force on the plant for a rightward horizontal saccade of the right eye is delivered by the pull of the $m_{LR}$, together with an equally rapid relaxation of the $m_{MR}$, while the other four tendons stay close to their equilibrium pretension values, albeit that the slight changes in tension during the horizontal saccades are systematic (cf. panels top-left and bottom-left). Similarly, an upward saccade requires the joint activation of the $m_{SR},~m_{IO}$, and a simultaneous inactivation of the $m_{IR},~m_{SO}$ muscles and vice versa for the downward saccade, with no change in net tension for the horizontal recti.

			\subsection{Discussion}
			
			\subsubsection{Summary} 
			We constructed a physics-based model for a biomimetic robotic eye controlled by six independent motors, each pulling an elastic string attached to the eye, and allowing it to rotate around its fixed center with three degrees of freedom (Fig.~\ref{fig:datasetdistribution}). String attachments resembled those of the human eye (Fig.~\ref{fig:eye}). The system dynamics were either approximated by analytical local linearization of the 3D equations of motion, or by learning the full nonlinear system properties by a neural network. We subjected both approximations to the same optimal control strategy that aimed to minimize three costs: 3D accuracy, movement duration, and the total energy expenditure of the motors. Our study demonstrates that both approximations yielded similar results, and that the control of the highly redundant oculomotor plant learned to generate rapid goal-directed eye rotations with human-like 3D kinematic and dynamic properties: eye movements followed nearly straight trajectories in all directions from any initial eye orientation within the oculomotor range (Fig.~\ref{fig:xyplanes}); movement duration increased with movement amplitude, and the peak velocity saturated with movement amplitude in a way that closely resembles the main-sequence properties of human saccades (Fig.~\ref{fig:eye_movement_properties} ); saccade trajectories closely followed Listing's law (Fig.~\ref{fig:xyplanes}), and the six motors resulted to organise themselves into antogonistic muscle pairs (Figs.~\ref{fig:motoranglesL} and \ref{fig:motoranglesNL}), much like its neurobiological counterpart. In what follows, we discuss these properties in more detail.
			
			\subsubsection{Linear vs. nonlinear approximations and pretension}
			Our results indicate that despite the significant differences in the linear vs. nonlinear approximations, the resulting control signals and eye-movement properties were quite similar. A potential disadvantage of the linearized model is the need to precalculate the pretension at every orientation in the work space, to avoid non-differentiable discontinuities for the local derivatives. In contrast, for the nonlinear NARX model, one has to set the pretensions for the central equilibrium orientation only, and through trial and error we selected relatively high values that guaranteed a convex total cost function (Fig. ~\ref{fig:sacc_cost_motor}). 
			
			Pretension in the tendons corresponds to a static co-contraction of the eye muscles, which is reportedly low in the primate oculomotor system as the eye has no stretch reflex \cite{robinson2022}\cite{robinson75}. Yet, the pools of oculomotor neurons for the agonist and antagonist eye muscles (as well as for the muscles not involved in the saccade) carry net neural activity for virtually all static eye positions \cite{robinson2022}\cite{hepp85}\cite{fuchs1970}\cite{robinson1972}. This indicates that although the eye doesn't have to deal with unexpected changes in load (unlike the skeletal muscles), or with the force of gravity, this net activity may serve a purpose. Possibly, it enables the system to exert fine control of the eye orientation by modulating firing rates, and to readily overcome static frictional forces that would induce unwanted hysteresis \cite{robinson2022}. Hence, the push-pull organization of the extraocular muscles with some static pretension may render the system with a high angular resolution.
			
			Although both models produced quite similar results, we believe that the nonlinear NARX approximation will be more flexible for potential future changes and added complexities, for which analytical derivatives may become increasingly more cumbersome. In addition, the NARX model could also be used to directly learn the control of the real robotic eye, without the need for explicit mathematical approximations of its physics.
			
			\subsubsection{Straight trajectories}
			
			Eye-movement trajectories in the $yz$ plane were relatively straight, despite the different dynamics for the horizontal and vertical saccade components (Fig.~\ref{fig:xyplanes}), and independent control of the six motors. A straight saccade implies that the horizontal and vertical velocity profiles are highly correlated (\cite{Gisb85}), and that the shape of the velocity profile of one component depends strongly on the presence of the other. Thus, the control signals for the six motors should be strongly cross-coupled, which will also depend on initial eye orientation and the goal direction and amplitude.
			
			Saccades generated by the linearized model were slightly straighter than those from the nonlinear NARX model. Yet, differences were small and most obvious for far-eccentric eye orientations (Fig.~\ref{fig:xyplanes}). Because straightness of  saccade trajectories was not included as a separate cost, it is an emerging property of the optimal control. Especially the requirement to reach the goal in minimum time, $J_D$, promotes eye movements as single-axis rotations of the eye during the entire trajectory. In that case, the six motors together effectively act as a central 3D vectorial eye-movement generator. Such a control has also been proposed for primate saccades, and the midbrain superior colliculus has been implicated to fulfill this role \cite{Gisb85}. 
			
			\subsubsection{Nonlinear main-sequence dynamics}
			The nonlinear main-sequence behavior of saccades was also obtained for our biomimetic system (Fig.~\ref{fig:dur_amp}). The nonlinear dynamics mainly resulted from the optimal control that effectively implements a speed-accuracy trade-off strategy, rather than from the nonlinearities in the plant \cite{Akhil2021}\cite{shadmehr2012}\cite{harris2006main}. Indeed, in our model, the increase in movement duration with movement amplitude was already seen in the identical increase in pulse/anti-pulse durations of the agonist/antagonist motors (Figs.~\ref{fig:motoranglesL} and \ref{fig:motoranglesNL}).

			As mentioned in the Introduction (Fig.~\ref{fig:eye_movement_properties}e), also the \textit{skewness} of human saccade velocity profiles positively correlates with saccade amplitude. This property, however, was not observed in our simulations (Fig.~\ref{fig:velocityprofile}), as velocity profiles remained nearly symmetrical for all amplitudes. Theoretical work on feedforward optimal control has suggested that the observed skewness could be due to the optimization of saccade accuracy in the presence of multiplicative noise in the control signals \cite{shadmehr2012}\cite{harris2006main}. According to this idea, it would be advantageous for the system to start saccades at high acceleration (with a low probability of final inaccuracy) and terminate with a gradual deceleration to ensure an accurate landing on the target. Preliminary work in our lab (not shown here) has indeed supported this idea also for our 3D biomimetic eye \cite{Cardoso2019}.

			\subsubsection{3D kinematics}
			In the present study, Listing's law (Fig.\ref{fig:xyplanes}) emerged from minimizing a cost function involving accuracy, duration and energy. Interestingly, although the goal in the accuracy cost (\ref{eq::accuracy}) was explicitly constrained to zero torsion, the optimal control yielded a plane in the laboratory reference frame that was slightly tilted in the $xy$ plane. A small leftward rotation of 4-10 deg around the $z$-axis aligned the data with Listing's frame. Apparently, the slight asymmetry in the muscular geometry of our robotic eye (MR being shorter than LR) was incompatible for the optimal control to generate eye orientations in true Listing coordinates with zero torsion. A subtle hint for this asymmetry effect was already observable in the $xy$ projection of the training data (Fig.~\ref{fig:datasetdistribution}). In our previous work \cite{Akhil2021}, we showed that Listing's law could emerge from the optimal control by minimizing the total fixation force at each eye orientation, with the goal specified in 2D. The plane would systematically tilt in the $xz$ projection when the vertical/torsional  muscles were displaced asymmetrically along the $y$-direction. We therefore conjecture that the orientation of Listing's plane, and hence the direction of the true primary position, results partly from the particular geometrical arrangement of the muscle insertions on the eye and head, in combination with their relative lengths and elastic properties. 
			
			\subsubsection{Antagonistic organisation}
			Because the linear and nonlinear approaches both resulted in similar controls for the six muscles (Figs.~\ref{fig:motoranglesL} and \ref{fig:motoranglesNL}), the joint antagonistic dynamic activation patterns are an emerging property of the optimal control strategy of our model, which does not critically depend on the exact details of the approximations of its plant dynamics.
			
			A  horizontal rightward saccade from straight ahead of the biomimetic eye comes about by a fast contraction (pulse) of the LR muscle, and a synchronous equally rapid relaxation (antipulse) of the MR muscle (Fig.~\ref{fig:motoranglesL}), while the four other muscles maintained their pretension innervations. Thus, the saccade is generated with minimal, or little, extra co-contraction in the eye muscles. Likewise, a purely vertical upward saccade from straight ahead results from the rapid synchronous contractions of the SR and IO muscles, together with a rapid relaxation of the IR and SO muscles, with the horizontal muscles kept at the equilibrium pretension (Fig.~\ref{fig:motoranglesL}). In this way, the net torsional action of the vertical-oblique pairs is close to zero, resulting in a purely vertical, upward eye rotation. Interestingly, oblique saccades involve the complex antagonistic interactions of all six muscles (Fig.~\ref{fig:motoranglesNL}). 
			
			These emerging synergies nicely incorporate the individual pulling directions of each muscle \cite{robinson75} (Fig. \ref{fig:muscle_pull_directions}, like e.g., the downward pulling action of the SO (a muscle located on top of the eye ball, see Fig.~\ref{fig:eye}) and therefore involved in downward saccades, and the upward pulling action of the IO (located at the bottom of the globe), recruited for upward saccades \cite{robinson75}\cite{suzuki99}\cite{fuchs1971}. The optimal control thus automatically results in what is known as the 'push-pull' organization of the oculomotor and vestibular systems, as prescribed by  Sherrington's famous principle of reciprocal innervation \cite{sherringon1906}.
			
			The pulse-step (and antipulse-antistep) innervation patterns of the agonist and antagonist muscles closely mimic their neurobiological counterparts of recorded primate oculomotor neurons \cite{robinson75}\cite{hepp85}\cite{suzuki99}\cite{fuchs1970}\cite{robinson1972}\cite{fuchs1971}. This reflects the built-in property of our biomimetic eye Eq. (\ref{full eye torque}) that the entire system behaves as an overdamped (albeit nonlinear) filter because of velocity-dependent dynamic friction \cite{robinson64}. In principle, it is possible to perform systems identification on the input (total six DOF muscle innervations) - output (3D eye orientation) relation of the model to determine the 3D transfer characteristic of the biomimetic plant, as well as how its properties depend on eye orientation. It would further be interesting to determine the equal-innervation trajectories for each of the six muscles as function of 3D eye orientation across the oculomotor range \cite{robinson75}\cite{hepp85}\cite{suzuki99}. These more elaborate studies fall beyond the scope of the present paper, however, and will be topic for a follow-up study. 
			
			\subsubsection{Current limitations and further study}
			Although the simulated model (Fig.~\ref{fig:graphical}) generated quite realistic human-like saccadic behaviors and neurobiologically plausible control signals, several aspects are simplified approximations of the actual robotic system of Fig.~\ref{fig:eyeprototype}, and may be further improved by adding more neurobiological and physical realism. First, the tendons of our simulator are attached to a fixed point on the eye ball, but follow the shortest, straight, path to the head-fixed insertion points at the motors (Fig.~\ref{fig:graphical}). Thus, a tendon's path may sometimes intersect with, and even pass through, the peripheral rim of the eye. An improved description would incorporate that each tendon wraps around the globe and leaves the eye at a tangent point that slides to a new eye-centered location as function of 3D eye orientation \cite{robinson75}. Moreover, in the primate eye, a portion of the muscle trajectory is fixed to the eye's sclera, which limits the amount of potential side-slip on the eye \cite{lee07}. The effective pulling direction of each eye muscle is thus determined by the direction of the vector pointing from the tangent point to the head-fixed insertion point, which has been suggested to be further modified by the presence of a pulley (e.g., \cite{demer06}, but see \cite{lee07}). Moreover, eye muscles consist of multiple fibers, rather than a single tendon as in our current implementation (Fig.~\ref{fig:eye}). Although we expect that including these additional properties will not qualitatively affect the current findings, they will influence the quantitative iso-innervation trajectories of the muscles \cite{robinson75}\cite{hepp85}, and the 3D plant characteristics.
			
			Second, in our current simulations we have not yet included the influence of additive and multiplicative noise on the control. It has been shown that especially the latter has an impact on the skewed shape of the velocity profiles (see above, and \cite{harris2006main}\cite{Cardoso2019}), and that it can replace the cost for energy expenditure as the accuracy cost with noise will contain the same quadratic control term \cite{shadmehr2012}.
			
			Finally, we here imposed a 3D constraint on movement accuracy, which implied that Listing's law ($r_x = 0$) was implicitly included as part of the accuracy cost, and that the primary position coincided with the head-fixed straight-ahead direction in the laboratory. In our previous work \cite{Akhil2021}, we included a quadratic cost for the total force on the eye, aiming to minimize fixation effort for peripheral eye orientations. In that case, a 2D accuracy cost sufficed to yield Listing's law, and that the primary position related to the muscular geometry. In the current study, with six independent motors, we had to specify the amount of pretension to prevent the muscles from going slack during eye movements. Possibly, by designing a quadratic fixation cost, combined with the previous two points, the optimal control may generate Listing's law, the primary position, and at the same time specify the optimal amount of pretension for the extraocular muscles at each eye orientation. This topic, however, will be explored in our future work.
			\section{Conclusion}
			\label{sec:conclusion}
			Our biomimetic implementation with six independent muscles is closer to neurobiological realism and leads to realistic 3D saccadic gaze- and control behaviors. We developed a full 3D non-linear dynamical model of our cable-driven robotic eye and compared linear and nonlinear approximations for its open-loop optimal control. Both approximations led to human-like saccadic eye movements with the correct 3D kinematics and dynamics in all directions. As the nonlinear neural-network approach does not critically depend on the details of the model to be learned, future improvements that introduce more complexity and realism to the system are better approximated by this method, as the linearized analytic approximations will become increasingly tedious. The NARX method can be readily implemented to learn the control dynamics for the true physical biomimetic eye, without the need to specify its physical properties in detail. 
			
			\section*{Acknowledgment}
			This research was supported by the European Union's Horizon 2020 program, ERC Advanced Grant, 2016 (project ORIENT, grant nr. 693400).
			%
			\ifCLASSOPTIONcaptionsoff
			\newpage
			\fi
			
			
			\bibliographystyle{IEEEtran}
			
			%
			%
			
			\bibliography{MyReferences} 
			
			\vspace{-14mm}
			
			
			

			\clearpage
			\appendices
			
			\section{Supplementary Materials} \label{sec:supplementary}

\begin{figure}[t]
	\centering
	\includegraphics[scale = 0.35]{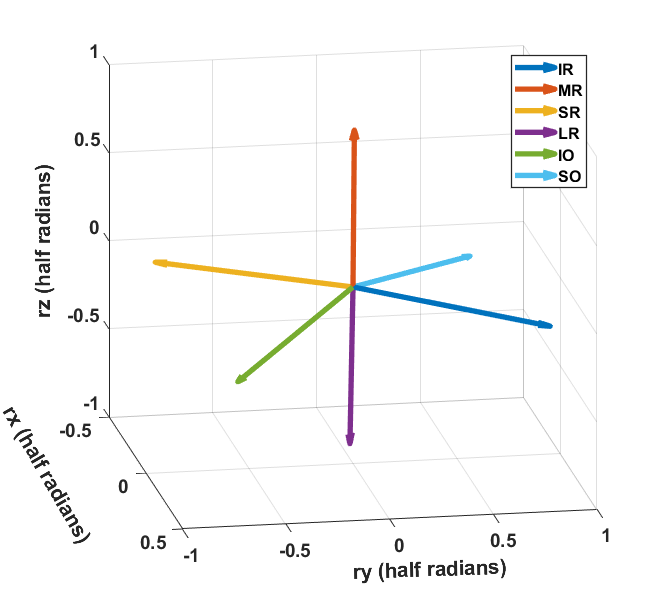}
	\caption{ Plot showing normalized direction of eye rotations (after 100 ms) when a single muscle is activated by rotating the corresponding motor by 0.1 radian while other muscles are held in pretension value. From the figure it's clear that the LR and MR muscles pull mostly in the horizontal direction (rz), while the other 4 muscles pull in both torsional and vertical directions.}
	\label{fig:muscle_pull_directions}
\end{figure}

\begin{figure}[t]
	\setlength\tabcolsep{1pt}
	\settowidth\rotheadsize{Radcliffe Cam}
	\setkeys{Gin}{width=\hsize}
	\begin{tabularx}{1\linewidth}{l| XXX }
		\rothead{\centering	\textcolor{blue}{Linear}}       
		&   \includegraphics[valign=m]{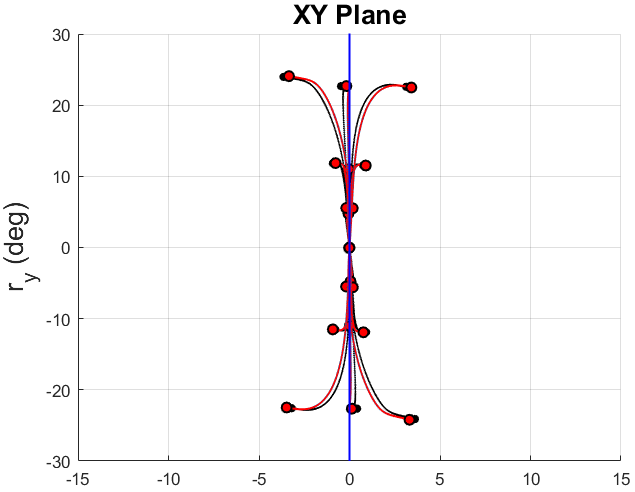}
		&   \includegraphics[valign=m]{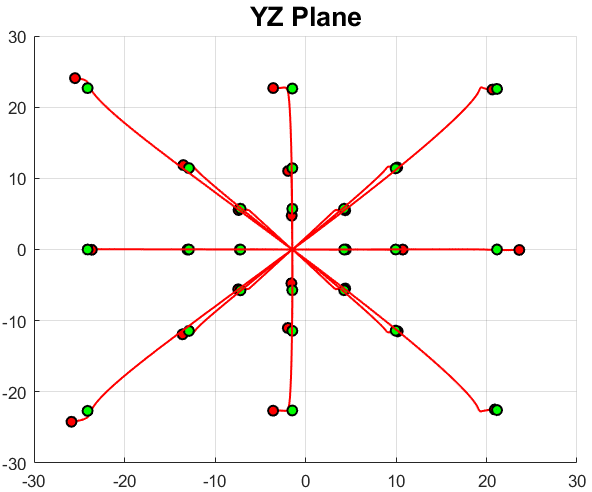}     \\  
		\addlinespace[2pt]
		\rothead{\centering \textcolor{blue}{Nonlinear}} 
		&   \includegraphics[valign=m]{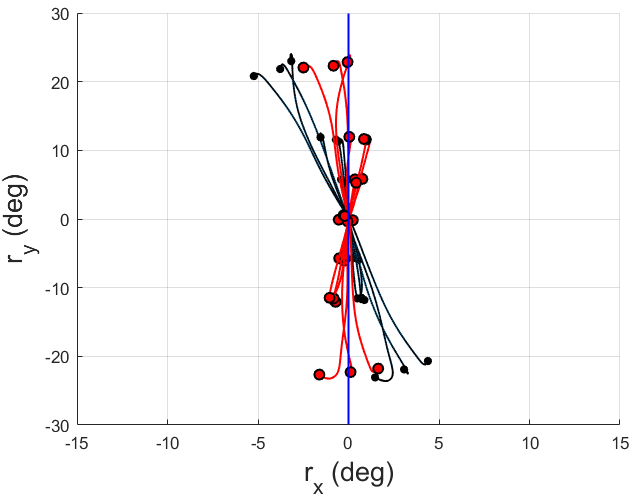}
		&   \includegraphics[valign=m]{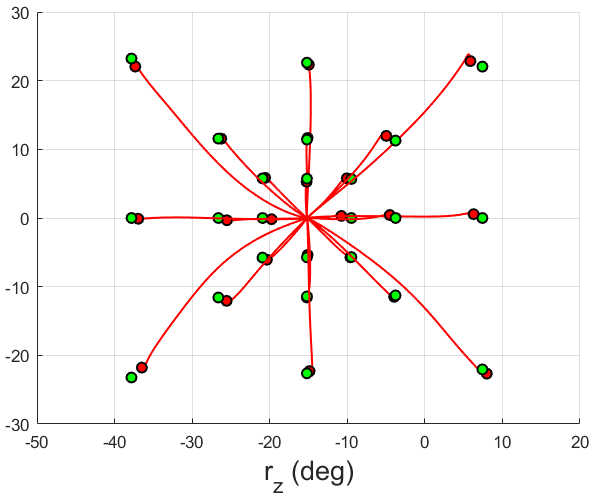}
	\end{tabularx}
	\caption{$xy$ (left) and $yz$ (right) projections of the instantaneous 3D eye-movement trajectories in Listing's coordinates (red) from the \textit{zero-initial} saccade sequence. Black: $xy$ data in laboratory frame before rotation about the $z$-axis into Listing's coordinates. \textbf{Top}: linearized approximation, rotation: 1.45 deg; \textbf{Bottom}: nonlinear approach, rotation: 15.2 deg. The blue vertical line in the $xy$ plots indicates Listing's plane; green dots correspond to the actual goals of the saccade set in Listing's frame.}
	\label{fig:xyplanes2}
\end{figure}

\begin{figure}
	\setlength\tabcolsep{1pt}
	\settowidth\rotheadsize{Radcliffe Cam}
	\setkeys{Gin}{width=\hsize}
	\begin{tabularx}{1\linewidth}{XX}
			\centering\textcolor{blue}{Linear} &  \textcolor{blue}{\qquad \qquad \quad Nonlinear}\\
	        \includegraphics[valign=m]{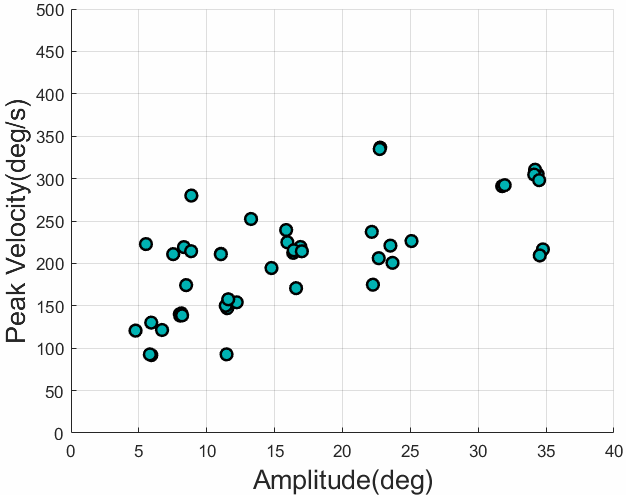} &  \includegraphics[valign=m]{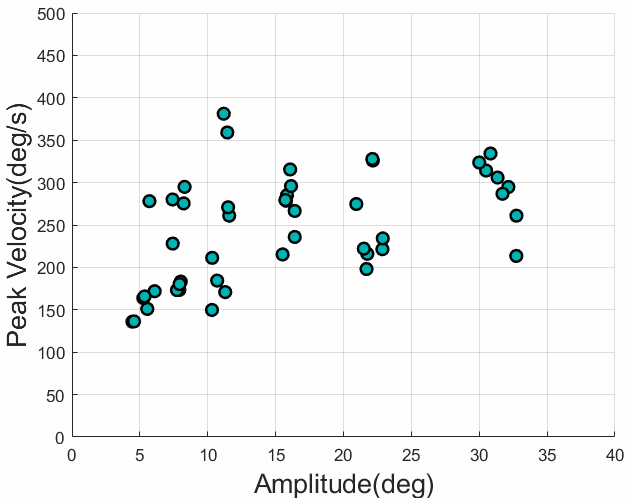}
	\end{tabularx}
	\caption{Amplitude - peak velocity relations for the linear and nonlinear approximations; all saccades pooled.
	}
	\label{fig:motoranglesNLAdditional}
\end{figure}

\begin{figure}
	\setlength\tabcolsep{1pt}
	\settowidth\rotheadsize{Radcliffe Cam}
	\setkeys{Gin}{width=\hsize}
	\begin{tabularx}{1\linewidth}{l XXX }
		\rothead{\centering	\textcolor{blue}{Linear}}       
		&   \includegraphics[valign=m]{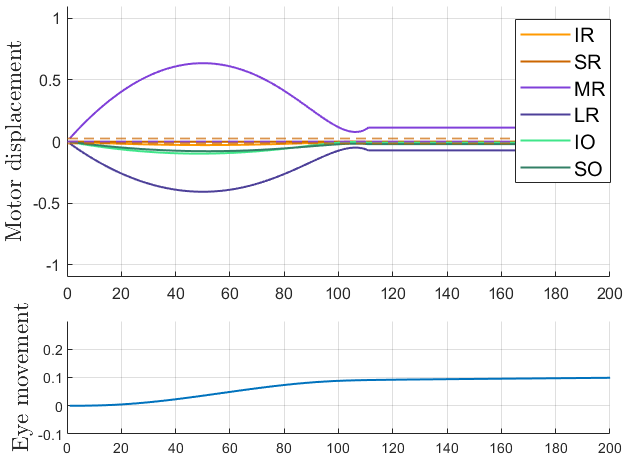}
		&   \includegraphics[valign=m]{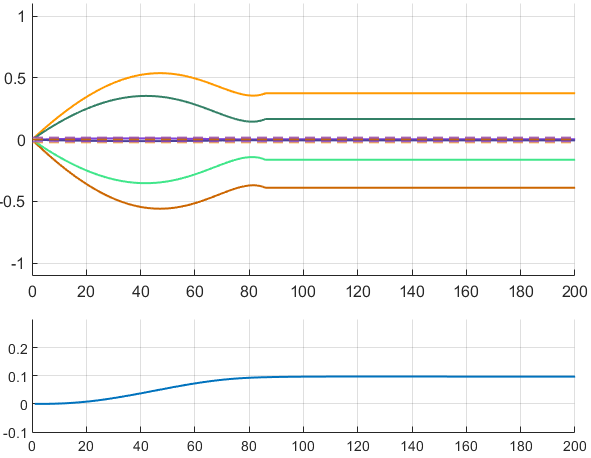}     \\  
		\addlinespace[2pt]
		\rothead{\centering \textcolor{blue}{Nonlinear}} 
		&   \includegraphics[valign=m]{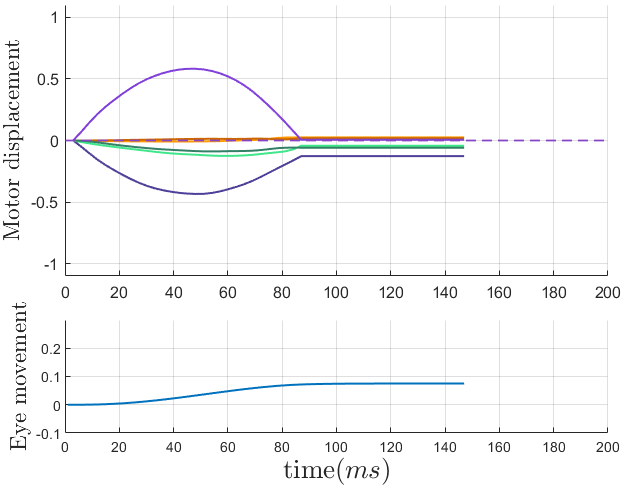}
		&   \includegraphics[valign=m]{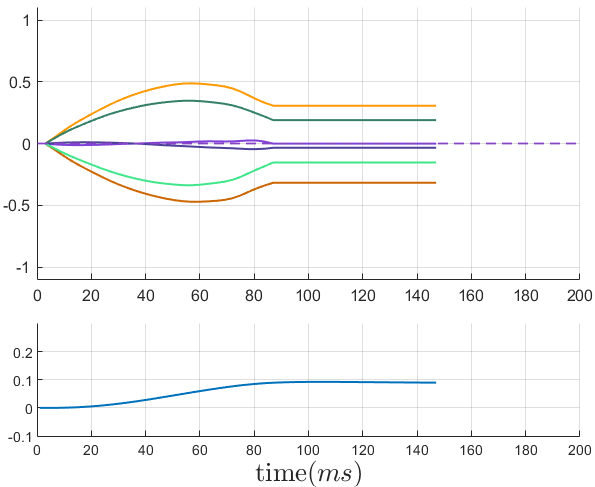}
	\end{tabularx}
		\caption{Changes in motor-command angles (radians) with respect to the pretension values for each tendon, associated with the optimal trajectories for two purely 11 deg rightward horizontal (left) and upward vertical (right) saccades, starting from the straight-ahead position with the linear and nonlinear controllers.
	}
	\label{fig:motoranglesAdditional}
\end{figure}

\begin{itemize}
    \item Fig. \ref{fig:muscle_pull_directions} that shows the rotation of the eye for each eye muscle, if only that muscle is activated by a small increment of the motor angle, with the others kept at their pretension.
    
    \item Fig. \ref{fig:xyplanes2} that shows all saccade trajectories (like in Fig. 10) for the central paradigm
    
    \item Fig. \ref{fig:motoranglesNLAdditional} that shows the amplitude-peak velocity relations for all saccades
    
     \item Fig. \ref{fig:motoranglesAdditional} that shows more examples of the pulse-step motor angles for oblique saccades in different directions and with different amplitudes.
    
\end{itemize}   
    

		\end{document}